\documentclass[english]{article}
\usepackage[T1]{fontenc}
\usepackage[utf8]{inputenc}
\usepackage{geometry}
\usepackage{color}
\usepackage{babel}
\usepackage{mathrsfs}
\usepackage{csquotes}
\usepackage{amsmath}

\usepackage{graphicx}
\usepackage{bm}
\usepackage{xcolor}
\usepackage{amssymb}
\usepackage{stmaryrd}
\usepackage{authblk}
\usepackage{subcaption}

\usepackage[backend=bibtex, style=numeric]{biblatex}
\begin{document}
\title{Physics-Consistent Neural Networks for Learning Deformation and Director Fields in Microstructured Media with Loss-Based Validation Criteria.}

\author[1]{Milad Shirani}
\author[1]{Pete H. Gueldner}
\author[2]{Murat Khidoyatov}
\author[1]{Jeremy L. Warren}
\author[1]{Federica Ninno}

\affil[1]{Department of Biomedical Engineering, Yale University, New Haven, CT, USA}
\affil[2]{Department of Mechanical Engineering, Yale University, New Haven, CT, USA}

\date{\today}
\maketitle

\begin{abstract}
In this work, we study the mechanical behavior of solids with microstructure using the framework of Cosserat elasticity with a single unit director. This formulation captures the coupling between deformation and orientational fields that arises in many structured materials. To compute equilibrium configurations of such media, we develop two complementary computational approaches: a finite element formulation based on variational principles and a neural network-based solver that directly minimizes the total potential energy. The neural architecture is constructed to respect the fundamental kinematic structure of the theory. In particular, it enforces frame invariance of the energy, satisfies the unit-length constraint on the director field, and represents deformation and director fields through separate networks to preserve their kinematic independence in the variational setting. Beyond satisfying balance laws, however, physically admissible solutions must also correspond to stable energy minimizers. To assess this requirement, we derive the quasiconvexity condition, rank-one convexity condition, and the Legendre-Hadamard inequalities for the Cosserat model and formulate them in a manner suitable for evaluating neural network predictions. These necessary stability conditions provide a physics-based validation framework: network outputs that violate these necessary conditions cannot correspond to stable energy minimizers and can therefore be rejected. In this way, we integrate classical variational stability theory with modern machine-learning solvers, establishing a computational workflow in which equilibrium solutions are not only learned but also assessed for energetic consistency.
\end{abstract}

\begin{flushleft}
\textbf{Keywords}: Physics-Informed Neural Network, Cosserat Elasticity, Quasiconvexity condition, Rank-one convexity condition, Legendre-Hadamard Inequalities.
\end{flushleft}

\section{Introduction} \label{Introduction}

Materials and biological systems with internal directional structure require continuum theories that couple deformation with local orientation fields. Several theories have been introduced to capture the effects of microstructure and these orientational properties into the modeling of the system. For example, higher gradient elasticity is used to model the effects of bending of the embedded fibers in tissues or fibrous materials \cite{shirani2025mixed, stilz2023second, stilz2024chirality, dell2019pantographic, placidi2017identification, spagnuolo2021higher, fedele2024review, fedele2022third, epstein2023kinematically}. Furthermore, Cosserat elasticity has been used to capture the effects of both bending and twisting of fibers in modeling fibrous materials \cite{shirani2025mixed, steigmann2015effects}. Cosserat elasticity with a single director is used to model mechanical behaviors of plates and shells \cite{green1968rods, steigmann1999relationship, eringen1997unified, green1996thermomechanical, green1976directed}, liquid crystals \cite{virga2018variational, ericksen2012theory, eringen1997unified, eringen1978micropolar, eringen1993assessment, ericksen1976equilibrium}, lipid bilayers and cell membranes \cite{steigmann2017role, shiraniLipidNormal, shiraniLipid, steigmannLiquidCrystalLipid1, steigmannLiquidCrystalLipid2, steigmannLiquidCrystalLipid3, steigmannLipidtilted1, steigmannLipidtilted2}, and cell monolayers \cite{yashunsky2022chiral, hoffmann2020chiral, chiang2024intercellular}.

A practical example of the usage of the theories of Cosserat elasticity with a single director is modeling the mechanical behaviors of nematic elastomers. In nematic elastomers, a polymer network's elastic response depends not only on deformation gradients but also on the orientation of internal anisotropic structures within the material, leading to distinctive mechanical behavior such as soft elasticity and anisotropic actuation that cannot be captured by classical elasticity alone \cite{xing2003fluctuating, xing2008nonlinear}. Similar continuum descriptions arise in liquid crystal systems and are relevant for structural materials where orientational order strongly influences mechanical response \cite{virga2018variational}. Beyond soft matter physics, Cosserat continuum models with internal fields are widely used in biomechanics and biological physics. For example, collective cell migration and polarization are modeled as continuum active media in which mechanical forces and internal polarity fields interact to govern tissue morphogenesis, wound healing, and cancer invasion \cite{banerjee2019continuum}.  By using theories of Cosserat continuum mechanics and liquid crystals, one can model the mechanical behaviors of lipid bilayers \cite{steigmann2017role}. By using these theories, one can show that the mechanical behaviors of lipid bilayers can be described by the mean and Gaussian curvatures of the body, leading to Helfrich energy \cite{steigmann2017role}. In these models, elastic bending and lateral stretching play key roles in membrane deformation, protein interactions, and cellular processes like endocytosis and vesicle formation \cite{kim2023review}. These examples illustrate how coupled deformation and internal structural fields can capture complex mechanical behavior across scales.

From a variational mechanics perspective, the existence and stability of solutions to elastic boundary-value problems hinge on properties of the energy density. In nonlinear elasticity, one can prove that choosing a convex functional form of elastic energy leads to unique solutions, and as a result, one cannot capture behaviors such as buckling \cite{steigmann2017finite}. Therefore, one has to find other conditions to prevent unrealistic cases coming from choosing convex elastic energy. It is shown that at the stable equilibrium, when the total potential energy of the system is minimized, the solution has to satisfy quasiconvexity, rank-one convexity, and Legendre-Hadamard inequalities. These conditions are necessary and must be satisfied by the solution to the balance laws and boundary conditions for them to be energy minimizers. In the seminal work of Ball \cite{ball1976convexity}, it is shown that the polyconvexity condition implies quasiconvexity conditions, rank-one convexity, and in consequence, Legendre-Hadamard inequalities. These conditions were established in classical analyses of nonlinear variational problems and elasticity theory, providing essential criteria for lower semicontinuity of energy functionals and absence of fine-scale instabilities \cite{ball1976convexity,dacorogna2008direct}. In continuum descriptions of structures such as plates and shells, similar energy minimization principles underlie generalized plate theories derived via variational asymptotics, such as those developed for active and composite thin sheets \cite{steigmann2023lecture,agostiniani2020rigorous}. These theoretical tools are central to assessing whether candidate solutions are not only stationary but stable under admissible perturbations, regardless of the specific physical system under study.

Recent work has extended these necessary stability conditions to continuum theories with additional internal fields. In particular, a Legendre-Hadamard condition appropriate to generalized continuum formulations that include structural fields, showing how positivity of the second variation imposes pointwise constraints on constitutive energy densities for stable minimizers \cite{shirani2020legendre}, and further work has explored rank-one convexity and quasiconvexity in related theories with additional field variables \cite{shirani2022quasiconvexity}. These developments provide robust theoretical checks that go beyond satisfying balance laws alone, ensuring that solutions are consistent with both mechanical equilibrium and energy minimization.

A variety of numerical approaches are used to approximate solutions to continuum mechanics problems with coupled fields. Traditional discretization methods such as finite element analysis remain the standard for many engineering and biophysical applications, including membrane deformation and thin structure mechanics. Emerging machine learning-based methods also embed physical laws directly into network training to approximate solutions to governing equations, and frameworks such as physics-informed neural networks have been applied in elasticity and related fields to solve for displacement and derived stress fields consistent with physical constraints \cite{raissi2019physics}. 

Physics-Informed Neural Networks (PINNs) for problems of classical continuum mechanics, such as nonlinear elasticity, were originally proposed to obtain the state of the system, by either solving balance laws \cite{raissi2019physics} or by minimizing the total potential energy of the system \cite{samaniego2020energy}. Related neural network approaches have also been used to predict tissue-level mechanical behavior and biological remodeling in vascular disease, including aneurysms, typically in a data-driven or hybrid modeling setting \cite{zhang2022g2varphinet, li2025importance, yin2022simulating, goswami2022neural, kontopodis2023prediction, chung2022artificial, chung2025augmented, rengarajan2021predictive, lindquist2021geometric}. Recent work suggests methods such as neural networks, Gaussian processes, ensembles, Support Vector Machines, regression, and PINNs can excel at fitting high-dimensional stress/strain mappings \cite{Hussain2024}. Supervised learning with feed-forward networks can directly learn stress-strain relationships, even outperforming the Holzapfel-Gasser-Ogden model with human aorta data \cite{Liu2020}. Similarly, the FHCNN-Bio model embeds fiber/invariant structure for improved interpretability, training stability, and strong generalization on experimental and synthetic tissue datasets \cite{Tan2025}. Symbolic regression approaches directly search for explicit, interpretable equations, using evolutionary/deep super-resolution (SR) to reduce model--form bias \cite{Abdusalamov2026}. PINNs learn constitutive behavior from limited or indirect measurements by enforcing PDEs/energy principles during training, showing promise for hyperelastic identification with sparse data \cite{Barrera2024}.

Even with such methods, however, satisfying balance equations is not sufficient: numerical approximations must be validated against necessary conditions for energy minimality to avoid spurious or unstable solutions. Despite these advances, a formal analysis of these approaches to check if the solutions satisfy energy--minimizing conditions has not been presented, mostly in non--classical continuum mechanics. However, some references \cite{chen2022polyconvex, klein2022polyconvex} have introduced networks for nonlinear classical elasticity that generate a polyconvex elastic energy. Note that, as shown by Ball \cite{ball1976convexity, steigmann2017finite}, if the elastic energy is polyconvex, then the solutions of the system automatically satisfy quasiconvexity and in turn rank--one convexity condition and the Legendre--Hadamard inequality.

In this manuscript, we formulate the governing balance laws and derive necessary variational stability criteria for a Cosserat continuum with a unit director, extending classical energy minimization principles to this class of coupled field problems. We then show how these criteria can be evaluated and used as validation checks for numerical solutions obtained from both finite element simulations and neural network--based solvers, ensuring consistency with energy minimality in addition to mechanical equilibrium. By integrating theory with computational approaches, our work provides a unified framework for assessing the mechanical and energetic consistency of solutions across a wide range of directed media, with applicability to systems ranging from soft biological tissues and active gels to thin structural elements and liquid crystalline materials.

\section{Cosserat mechanics of microstructured media} \label{Kinematics and elastic energy}

In this section, we present the mechanical modeling of materials by including the effects of their microstructures on their mechanical behaviors. We use the theory of Cosserat elasticity with a unit director to model these media. 

\subsection{Kinematics and frame invariance condition}
In this section, we present the kinematics of a body described by the theories of Cosserat elasticity with a single unit director. According to this theory, the kinematics of the body is described by two kinematical
fields, namely deformation field from a reference configuration $\kappa_{0}$
to the current configuration $\kappa$ and a unit director field in
the current configuration. As a result, the kinematics of the motion
are
\begin{equation}
\bm{x}=\bm{\chi}\left(\bm{X}\right)\qquad\qquad\text{and}\qquad\qquad\bm{d}=\bm{d}\left(\bm{X}\right)
\end{equation}
in which $\bm{X}$ is a material point in the reference configuration
and $\bm{x}$ is the position of the same material point after motion
in the current configuration. In this work, we require $\bm{d}$ to
be a unit vector, i.e., 
\begin{equation}
\bm{d}\cdot\bm{d}=1\label{constraint}
\end{equation}
and this should be considered as a constraint in the formulation of
the problem. We assume that there exists an elastic energy shown by
$W$ defined per unit volume of the reference configuration
\begin{equation}
W=W\left(\bm{F},\bm{d},\nabla\bm{d}\right)\label{elastic energy}
\end{equation} 
in which $\bm{F}$ is the gradient of the deformation, defined as
\begin{equation}
\bm{F}=\frac{\partial\bm{\chi}}{\partial X_{A}}\otimes\bm{E}_{A}=\nabla\bm{\chi}
\end{equation}
in which $\bm{E}_{A}$ are the orthonormal basis vectors in the reference
configuration and $\nabla\bullet$ is the referential gradient of
the quantity $\bullet$. We require deformation $\bm{\chi}$ be invertible
and as a result, we find 
\begin{equation}
J=\det\bm{F}>0.
\end{equation}

Elastic energy has to be frame invariant, meaning that for any $\varepsilon-$family
of rotation tensors $\bm{Q}\left(\varepsilon\right)$, the following relation has to hold
\begin{equation}
W\left(\bm{F},\bm{d},\nabla\bm{d}\right)=W\left(\bm{Q}\left(\varepsilon\right)\bm{F},\bm{Q}\left(\varepsilon\right)\bm{d},\bm{Q}\left(\varepsilon\right)\nabla\bm{d}\right).\label{energy with rotation in objectivity}
\end{equation}
Given that the left--hand side of (\ref{energy with rotation in objectivity})
is not a function of $\varepsilon$, but the right--hand side is, we
conclude that
\begin{equation}
\frac{d}{d\varepsilon}\left(W\left(\bm{Q}\left(\varepsilon\right)\bm{F},\bm{Q}\left(\varepsilon\right)\bm{d},\bm{Q}\left(\varepsilon\right)\nabla\bm{d}\right)\right)=0
\end{equation}
and as a result
\begin{equation}
\left(\frac{\partial W}{\partial\left(\bm{Q}\left(\varepsilon\right)\bm{F}\right)}\bm{F}^{T}+\frac{\partial W}{\partial\left(\bm{Q}\left(\varepsilon\right)\bm{d}\right)}\otimes\bm{d}+\frac{\partial W}{\partial\left(\bm{Q}\left(\varepsilon\right)\nabla\bm{d}\right)}\left(\nabla\bm{d}\right)^{T}\right)\cdot\frac{d\bm{Q}\left(\varepsilon\right)}{d\varepsilon}=0
\end{equation}

Among all possible choices of rotation tensors, choose $\bm{Q}\left(\varepsilon\right)$
such that $\bm{Q}\left(0\right)=\bm{I}$ and $\left(\frac{d\bm{Q}\left(\varepsilon\right)}{d\varepsilon}\right)_{\mid\varepsilon=0}=\bm{\Omega}$, where $\bm{\Omega}$ is an arbitrary second order skew--symmetric tensor ($\bm{\Omega} \in \text{Skew}$).
By evaluating the above relation at $\varepsilon=0$, we get 
\begin{equation}
\left(\frac{\partial W}{\partial\bm{F}}\bm{F}^{T}+\frac{\partial W}{\partial\bm{d}}\otimes\bm{d}+\frac{\partial W}{\partial\nabla\bm{d}}\left(\nabla\bm{d}\right)^{T}\right)\cdot\bm{\Omega}=0
\end{equation}
and because $\bm{\Omega}$ is an arbitrary skew--symmetric tensor,
we conclude
\begin{equation}
\frac{\partial W}{\partial\bm{F}}\bm{F}^{T}+\frac{\partial W}{\partial\bm{d}}\otimes\bm{d}+\frac{\partial W}{\partial\nabla\bm{d}}\left(\nabla\bm{d}\right)^{T}\in\text{Sym}\label{Objectivity symmetric}
\end{equation}
Any energy used for the system has to satisfy (\ref{Objectivity symmetric}), otherwise, it is not a frame--invariant elastic energy. 

\subsection{Balance laws and boundary conditions} \label{Balance laws and boundary conditions}

At equilibrium, the total potential energy of the system is minimized.
As a result, one could conclude that the first variation of the total potential energy of the system has to vanish. Therefore, to find balance
laws and boundary conditions, we will use the principle of virtual
power and the calculus of variations. According to the principle of virtual
power, for any kinematically admissible virtual deformation and virtual
director, the variation of the elastic energy of the system is equal
to the virtual power of all applied effects. Note that, one has to
consider the constraints in the formulation of the problem. In this
work, the only constraint that we have is (\ref{constraint}). Therefore,
we need to find the variation of the director field such that this
constraint is satisfied. Consider one $\varepsilon-$family of the
variations of deformation and director fields as
\begin{align}
\hat{\bm{\chi}}\left(\bm{X},\varepsilon\right) & =\bm{\chi}\left(\bm{X}\right)+\varepsilon\delta\bm{\chi}+\bm{o}\left(\varepsilon\right)\\
\hat{\bm{d}}\left(\bm{X},\varepsilon\right) & =\bm{d}\left(\bm{X}\right)+\varepsilon\delta\bm{d}+\bm{o}\left(\varepsilon\right)
\end{align}
in which $\bm{\chi}$ and $\bm{d}$ are the kinematical field at the
equilibrium, $\delta\bm{\chi}$ and $\delta\bm{d}$ are the variation
of the deformation and the director field, respectively. Variation of the director field has to satisfy the constraint. Therefore, taking the variation of (\ref{constraint}) results in
\begin{equation}
\bm{d}\cdot\delta\bm{d}=0
\end{equation}
from which we conclude that $\delta\bm{d}=\bm{g}\times\bm{d}$ in
which $\bm{g}$ is a non--unique arbitrary vector. By using this, and showing
the variation of the deformation by $\bm{u}$, we can write
\begin{align}
\hat{\bm{\chi}}\left(\bm{X},\varepsilon\right) & =\bm{\chi}\left(\bm{X}\right)+\varepsilon\bm{u}+\bm{o}\left(\varepsilon\right)\\
\hat{\bm{d}}\left(\bm{X},\varepsilon\right) & =\bm{d}\left(\bm{X}\right)+\varepsilon\bm{g}\times\bm{d}+\bm{o}\left(\varepsilon\right)
\end{align}
and in consequence, we find $\delta\bm{F}$ and $\nabla\delta\bm{d}$
as
\begin{equation}
\delta\bm{F}=\nabla\bm{u}\qquad\qquad\text{and}\qquad\qquad\nabla\delta\bm{d}=\left(\bm{g}_{,A}\times\bm{d}+\bm{g}\times\bm{d}_{,A}\right)\otimes\bm{E}_{A}
\end{equation}

Taking the variation of the elastic energy (\ref{elastic energy})
gives
\begin{equation}
\delta W=\bm{P}\cdot\delta\bm{F}+\bm{N}\cdot\delta\bm{d}+\bm{M}\cdot\nabla\delta\bm{d}
\end{equation}
in which $\bm{P}$ is the generalized first Piola stress tensor, $\bm{N}$
is the internal couple, and $\bm{M}$ is couple stress, defined as
\begin{equation}
\bm{P}=\frac{\partial W}{\partial\bm{F}},\qquad\bm{N}=\frac{\partial W}{\partial\bm{d}},\qquad\bm{M}=\frac{\partial W}{\partial\nabla\bm{d}} \label{stresses}
\end{equation}

Taking the volume integral of $\delta W$ gives
\begin{align}
\int_{\kappa_{0}}\delta W\,dV & =\int_{\kappa_{0}}\left(\bm{P}\cdot\delta\bm{F}+\bm{N}\cdot\delta\bm{d}+\bm{M}\cdot\nabla\delta\bm{d}\right)dV\\
 & =\int_{\kappa_{0}}\left(\text{Div}\left(\bm{P}^{T}\bm{u}\right)-\text{Div}\left(\bm{P}\right)\cdot\bm{u}+\bm{d}\times\left(\bm{N}-\text{Div}\left(\bm{M}\right)\right)\cdot\bm{g}+\text{Div}\left(\bm{M}^{T}\delta\bm{d}\right)\right)dV
\end{align}
where $dV$ is the element of volume in the reference configuration
and by applying the divergence theorem for integrals, we find
\begin{align}
\int_{\kappa_{0}}\delta W\,dV & =\int_{\partial\kappa_{0}}\left(\bm{P}\bm{\nu}\cdot\bm{u}+\bm{d}\times\bm{M}\bm{\nu}\cdot\bm{g}\right)dA\nonumber -\int_{\kappa_{0}}\left(\text{Div}\left(\bm{P}\right)\cdot\bm{u}+\bm{d}\times\left(\text{Div}\left(\bm{M}\right)-\bm{N}\right)\cdot\bm{g}\right)dV
\end{align}
in which $dA$ is the referential areal element, $\bm{\nu}$ is the outward unit normal to the boundary of the body in the reference configuration, $\partial\kappa_{0}$ is the boundary of the body in the reference configuration. According to the principle of virtual power, at equilibrium,
we have
\begin{equation}
\int_{\kappa_{0}}\delta W\,dV=\mathcal{P}
\end{equation}
in which $\mathcal{P}$ is the virtual power of external effects, i.e., 
\begin{equation}
\mathcal{P}=\int_{\kappa_{_{0}}}\left(\bm{f}\cdot\bm{u}+\bm{l}\cdot\bm{g}\times\bm{d}\right)\,dV+\int_{\partial\kappa_{p_{0}}}\bm{p}\cdot\bm{u}\,dA+\int_{\partial\kappa_{d_{0}}}\bm{m}\cdot\bm{g}\times\bm{d}\,dA
\end{equation} 
in which $\bm{f}$, $\bm{l}$, $\bm{p}$, and $\bm{m}$ are external
body force, body couple, traction, and couple traction, respectively,
and $\partial\kappa_{p_{0}}$ and $\partial\kappa_{d_{0}}$ are portion
of the boundary of the body over which $\bm{p}$ and $\bm{m}$ are
applied. Note that over the parts of the boundary where deformation
and director fields are, which prescribe the variations of the deformation
and director fields vanish. By applying the principle of virtual power,
one can find the following balance laws and boundary conditions
\begin{align}
\text{Div}\bm{P}+\bm{f} & =\bm{0}\quad\in\kappa_{0}\qquad\qquad\text{and}\qquad\qquad\bm{P}\bm{\nu}=\bm{p}\qquad\text{on }\partial\kappa_{p_{0}}\label{Balance of Linear Momentum}\\
\bm{d}\times\left(\text{Div}\left(\bm{M}\right)-\bm{N}+\bm{l}\right) & =\bm{0}\quad\in\kappa_{0}\qquad\qquad\text{and}\qquad\qquad\bm{d}\times\bm{M}\bm{\nu}=\bm{d}\times\bm{m}\qquad\text{on }\partial\kappa_{d_{0}}\label{Balance of Director Momentum}
\end{align}

Given that in this work we are proposing neural network methods to model the effects of materials' microstructure on the mechanical behavior of the body of the body based on variational methods, in the following sections, we will present conditions that must be satisfied by the outputs of the network for any form of elastic energy used as part of the loss function in the network. We obtain quasiconvexity condition, rank--one convexity condition, and Legendre--Hadamard inequalities that must be satisfied by the solutions of these balance laws and boundary conditions for them to correspond to stable energy minimizers.

\subsection{Energy minimizers} \label{Energy minimizers}

In this section, we derive energy--minimizing conditions that the solutions of (\ref{Balance of Linear Momentum}) and (\ref{Balance of Director Momentum}) have to satisfy. Therefore, these conditions are necessary for these solutions, and if the solutions do not satisfy these necessary conditions, they cannot correspond to stable minimizers. These conditions are important because they provide a ground truth to check the validity of a solution obtained by numerical methods or analytical approaches for conservative systems at equilibrium. As a result, we are using the total potential energy of the system as a loss function in neural networks to make sure that the solutions of the networks adhere to and satisfy the laws of physics and thermodynamics. We conclude that these energy--minimizing conditions also must be satisfied by the output of the networks for any given loss function. Therefore, if the solutions obtained by the neural networks do not satisfy these conditions, we conclude that the network's solutions are not valid.

At the stable equilibrium, the total potential energy of a system
is minimized. Consider the total potential energy of a conservative system as
\begin{equation}
\mathcal{E}\left(\bm{\chi},\bm{d}\right)=\int_{\kappa_{0}}W\left(\bm{F},\bm{d},\nabla\bm{d}\right)dV-\mathcal{L}\label{total potential energy}
\end{equation}
in which $\mathcal{L}$ is the potential energy of applied external
effect. In this work, we consider the case of dead loads and in consequence
\begin{equation}
\mathcal{L}=\int_{\kappa_{_{0}}}\left(\bm{f}\cdot\bm{\chi}+\bm{l}\cdot\bm{d}\right)\,dV+\int_{\partial\kappa_{p_{0}}}\bm{p}\cdot\bm{\chi}\,dA+\int_{\partial\kappa_{d_{0}}}\bm{m}\cdot\bm{d}\,dA 
\end{equation}

For any deformation and director fields satisfying boundary conditions
and that are not the kinematical fields at equilibrium, the total
potential energy of the system should be greater than the value of
the total potential energy for the kinematical fields at the equilibrium.
That is to say
\begin{equation}
\mathcal{E}\left(\tilde{\bm{\chi}},\tilde{\bm{d}}\right)-\mathcal{E}\left(\bm{\chi}_{e},\bm{d}_{e}\right)\geq0\label{Difference in potential energy}
\end{equation}
in which $\bm{\chi}_{e}$ and $\bm{d}_{e}$ are the kinematical fields
at the stable equilibrium obtained from solving (\ref{Balance of Linear Momentum})
and (\ref{Balance of Director Momentum}) whereas and $\tilde{\bm{\chi}}$
and $\tilde{\bm{d}}$ are the deviated kinematical fields away from
equilibrium. Note that both $\tilde{\bm{d}}$ and $\bm{d}_{e}$ have
to satisfy the constraint in (\ref{constraint}). In the following sections,
we find quasiconvexity condition, rank--one convexity condition, and
Legendre--Hadamard inequalities by minimizing the total potential energy
of the system. These conditions must be satisfied by the solutions
of (\ref{Balance of Linear Momentum}) and (\ref{Balance of Director Momentum})
at the stable equilibrium. Therefore, these conditions are necessary
conditions that must be satisfied by solutions of the balance laws
for them to be valid at the equilibrium. 

\subsubsection{Quasiconvexity condition} \label{Quasiconvexity condition}

Consider the following deformations
\begin{equation}
\tilde{\bm{\chi}}\left(\bm{X},\mu\right)=\bm{\chi}_{e}\left(\bm{X}\right)+\mu\bm{\Phi}\left(\bm{Z}\right)\qquad\text{and}\qquad\tilde{\bm{d}}\left(\bm{X},\mu\right)=\bm{R}\left(\bm{Z},\mu\right)\bm{d}_{e}\qquad\text{with}\qquad\bm{R}\left(\bm{Z},\mu\right)\in\text{Orth}^{+}\label{compactly supported deformations}
\end{equation}
in which $\bm{\Phi}\left(\bm{Z}\right)$ and $\tilde{\bm{R}}\left(\bm{Z},\mu\right)-\bm{I}$
are smooth enough compactly supported functions in $\mathcal{D}\subset\kappa$
and 
\begin{equation}
\bm{Z}=\frac{\bm{X}-\bm{X}_{0}}{\mu}
\end{equation}
where $\bm{X}_{0}$ is a material point in the interior of $\kappa_0$
and $\mu$ is a scalar. Note that choosing $\tilde{\bm{d}}\left(\bm{X},\mu\right)=\tilde{\bm{R}}\left(\bm{Z},\mu\right)\bm{d}_e$
satisfies the constraint (\ref{constraint}) therefore, it is a valid
director field. There are several representations for a rotation tensor,
among which we use Cayley's representation of rotation tensors. According
to this representation, for any skew--symmetric tensor $\bm{W}$ the
quantity $\left(\bm{I}-\bm{W}\right)^{-1}\left(\bm{I}+\bm{W}\right)$
is a rotation tensor. Therefore, by using this representation, we
pick 
\begin{equation}
\bm{R}\left(\bm{Z},\mu\right)=\left(\bm{I}-\frac{\mu}{2}\bm{W}\left(\bm{Z}\right)\right)^{-1}\left(\bm{I}+\frac{\mu}{2}\bm{W}\left(\bm{Z}\right)\right)\label{Cayley Representation}
\end{equation}
in which $\bm{W}\left(\bm{Z}\right)$ is a smooth compactly supported 
tensor--valued function in $\mathcal{D}\subset\kappa_{0}$. By using
(\ref{compactly supported deformations}) and (\ref{Cayley Representation}),
we find
\begin{align}
\tilde{\bm{F}} & =\bm{F}_{e}+\nabla_{_{\bm{Z}}}\bm{\Phi}\label{tilde F quasi convexity}\\
\nabla\tilde{\bm{d}} & =\frac{1}{\mu}\frac{\partial\bm{R}}{\partial Z_{A}}\bm{d}_{e}\otimes\bm{E}_{A}+\bm{R}\nabla\bm{d}_{e} \label{tilde grad d for quasi convexity}
\end{align}
in which
\begin{equation}
\nabla_{_{\bm{Z}}}\bullet=\frac{\partial\bullet}{\partial Z_{A}}\otimes\bm{E}_{A}
\end{equation}
 is the gradient of the vector quantity $\bullet$. Note that
\begin{equation}
\frac{1}{\mu}\frac{\partial\bm{R}}{\partial Z_{A}}=\frac{1}{2}\left(\bm{I}-\frac{\mu}{2}\bm{W}\right)^{-1}\left(\frac{\partial\bm{W}}{\partial Z_{A}}\right)\left(\bm{R}+\bm{I}\right) \label{gradient of R}
\end{equation}
where we used the identity
\begin{equation}
\frac{\partial}{\partial Z_{A}}\left(\bm{I}-\frac{\mu}{2}\bm{W}\right)^{-1}=\left(\bm{I}-\frac{\mu}{2}\bm{W}\right)^{-1}\left(\frac{\mu}{2}\frac{\partial\bm{W}}{\partial Z_{A}}\right)\left(\bm{I}-\frac{\mu}{2}\bm{W}\right)^{-1} \label{gradient of the (I - W)-1}
\end{equation}
and by substituting (\ref{gradient of the (I - W)-1}) into (\ref{gradient of R}) and results in (\ref{tilde grad d for quasi convexity}), we find $\nabla \tilde{\bm{d}}$ as
\begin{equation}
\nabla\tilde{\bm{d}}=\frac{1}{2}\left(\bm{I}-\frac{\mu}{2}\bm{W}\right)^{-1}\left(\frac{\partial\bm{W}}{\partial Z_{A}}\right)\left(\bm{R}+\bm{I}\right)\bm{d}_{e}\otimes\bm{E}_{A}+\bm{R}\nabla\bm{d}_{e}\label{nabla tilde d quasi convexity} 
\end{equation}

By substituting (\ref{compactly supported deformations}), (\ref{tilde F quasi convexity}),
and (\ref{nabla tilde d quasi convexity}) into (\ref{Difference in potential energy})
and using the fact that $\bm{\Phi}=\bm{0}$ on the boundary $\partial\kappa_{p_{0}}$
and $\bm{R}=\bm{I}$ on $\partial\kappa_{d_{0}}$, we find
\begin{align}
\mathcal{E}\left(\tilde{\bm{\chi}},\tilde{\bm{d}}\right)-\mathcal{E}\left(\bm{\chi}_{e},\bm{d}_{e}\right) & =\int_{\mathcal{D}}W\left(\bm{F}_{e}+\nabla_{_{\bm{Z}}}\bm{\Phi},\bm{R}\bm{d}_{e},\bm{R}\nabla\bm{d}_{e}+\frac{1}{2}\left(\bm{I}-\frac{\mu}{2}\bm{W}\right)^{-1}\left(\frac{\partial\bm{W}}{\partial Z_{A}}\right)\left(\bm{R}+\bm{I}\right)\bm{d}_{e}\otimes\bm{E}_{A}\right)\mu^{3}dV_{Z}\nonumber \\
 & -\int_{\mathcal{D}}W\left(\bm{F}_{e},\bm{d}_{e},\nabla\bm{d}_{e}\right)\mu^{3}dV_{Z}-\int_{\mathcal{D}}\left(\bm{f}\cdot\mu\bm{\Phi}\left(\bm{Z}\right)+\bm{l}\cdot\left(\bm{R}-\bm{I}\right)\bm{d}_{e}\right)\,\mu^{3}dV_{Z}\geq0
\end{align}
in which $dV_{Z}$ is the volume element of the domain $\mathcal{D}$.
By dividing this relation by $\mu^{3}$, passing to the limit $\mu\to0$,
and because $\lim_{\mu\to0}\bm{R}=\bm{I}$, and $\lim_{\mu\to0}\bm{X}=\bm{X}_{0}$,
we find 
\begin{equation}
\int_{\mathcal{D}}\left[W\left(\bm{F}_{0e}+\nabla_{_{\bm{Z}}}\bm{\Phi},\bm{d}_{0e},\nabla\bm{d}_{0e}+\frac{\partial\bm{W}}{\partial Z_{A}}\bm{d}_{0e}\otimes\bm{E}_{A}\right)-W\left(\bm{F}_{0e},\bm{d}_{0e},\nabla\bm{d}_{0e}\right)\right]dV_{Z}\geq0\label{quasiconvexity}
\end{equation}
in which the quantities with the subscription $0e$ are those at the
equilibrium and they are evaluated at $\bm{X}=\bm{X}_{0}.$ Relation
(\ref{quasiconvexity}) is called quasiconvexity condition which is
a necessary condition for $\bm{d}_{e}$ and $\bm{\chi}_{e}$ to be
those at the stable equilibrium. This relation has to hold for any
compactly supported functions $\bm{\Phi}$ and $\bm{W}$ and for any
domain $\mathcal{D}\subset\kappa_{0}$ and at every material point
$\bm{X}_{0}$. 

Because $\bm{W}$ is a skew symmetric tensor, we use its axial vector
$\bm{w}=\text{ax}\left(\bm{W}\right)$ and rewrite the above relation
as
\begin{equation}
\int_{\mathcal{D}}\left[W\left(\bm{F}_{0e}+\nabla_{_{\bm{Z}}}\bm{\Phi},\bm{d}_{0e},\nabla\bm{d}_{0e}+\frac{\partial\bm{w}}{\partial Z_{A}}\times\bm{d}_{0e}\otimes\bm{E}_{A}\right)-W\left(\bm{F}_{0e},\bm{d}_{0e},\nabla\bm{d}_{0e}\right)\right]dV_{Z}\geq0.\label{Quasiconvexity with axial vector} 
\end{equation}

It can be seen that the quasiconvexity condition is an integral inequality that
has to hold for any domain $\mathcal{D}$, for any compactly supported
functions $\bm{\Phi}$ and $\bm{W}$ at each material point. Therefore,
one can conclude that it would not be possible to use it to check
the validity of the solutions. In the next section, we obtain rank--one
convexity condition, which is a point--wise condition that must be satisfied
by $\bm{\chi}_{e}$ and $\bm{d}_{e}$. 

\subsubsection{Rank--one convexity condition} \label{Rank-one convexity condition}

In this section, we obtain rank--one convexity condition for Cosserat
media with a single unit director field. The origin of $\bm{Z}$ is
located at $\bm{X}=\bm{X}_{0}$, therefore, by a coordinate transformation
and attaching an orthonormal right--handed basis $\left\{ \bm{L}_{1},\bm{L}_{2},\bm{L}_{3}\right\} $
on point $\bm{X}_{0}$, we can define a new coordinate system $\left(\xi_{1},\xi_{2},\xi_{3}\right)$
such that $\xi_{1}=\bm{Z}\cdot\bm{L}_{1}$, $\xi_{2}=\bm{Z}\cdot\bm{L}_{2}$,
and $\xi_{3}=\bm{Z}\cdot\bm{L}_{3}$; therefore, we can write $\bm{Z}$
as
\begin{equation}
\bm{Z}=\xi_{1}\bm{L}_{1}+\xi_{2}\bm{L}_{2}+\xi_{3}\bm{L}_{3}
\end{equation}
Note that given that $\left\{ \bm{L}_{A}\right\} $ and $\left\{ \bm{E}_{A}\right\} $
are orthonormal basis vectors, there exists a rotation tensor $\bm{\mathcal{S}}$
such that $\bm{E}_{A}=\bm{\mathcal{S}}\bm{L}_{A}$. Moreover, consider
a domain $\mathcal{R}\subset\mathcal{D}$ such that
\begin{equation}
\mathcal{R}=\left\{ \left(\xi_{1},\xi_{2},\xi_{3}\right)|\left|\xi_{3}\right|<B\left(\xi_{1},\xi_{2}\right)=h^{2}-\left(\xi_{1}^{2}+\xi_{2}^{2}\right)\qquad\text{and}\qquad B\left(\xi_{1},\xi_{2}\right)\geq0\right\} 
\end{equation}
where $h$ is small enough. Consider disjoint domains $\mathcal{R}_{0},$
$\mathcal{R}_{+}$ and $\mathcal{R}_{-}$ such that $\mathcal{R}=\mathcal{R}_{-}\cup\mathcal{R}_{0}\cup\mathcal{R}_{+}$
\begin{align}
\mathcal{R}_{+} & =\left\{ \left(\xi_{1},\xi_{2},\xi_{3}\right)\qquad|\qquad\theta\,B\left(\xi_{1},\xi_{2}\right)<\xi_{3}\leq B\left(\xi_{1},\xi_{2}\right)\right\} \\
\mathcal{R}_{0} & =\left\{ \left(\xi_{1},\xi_{2},\xi_{3}\right)\qquad|\qquad-\theta\,B\left(\xi_{1},\xi_{2}\right)\leq\xi_{3}\leq\theta\,B\left(\xi_{1},\xi_{2}\right)\right\} \\
\mathcal{R}_{-} & =\left\{ \left(\xi_{1},\xi_{2},\xi_{3}\right)\qquad|\qquad-B\left(\xi_{1},\xi_{2}\right)\leq\xi_{3}<-\theta\,B\left(\xi_{1},\xi_{2}\right)\right\} 
\end{align}
with $0<\theta\leq1$. Consider the following function
\begin{equation}
f\left(\bm{Z}\right)=\begin{cases}
-\theta\left[\xi_{3}-B\left(\xi_{1},\xi_{2}\right)\right] & ,\bm{Z}\in\mathcal{R}_{+}\\
\left(1-\theta\right)\xi_{3} & ,\bm{Z}\in\mathcal{R}_{0}\\
-\theta\left[\xi_{3}+B\left(\xi_{1},\xi_{2}\right)\right] & ,\bm{Z}\in\mathcal{R}_{-}\\
0 & ,\bm{Z}\notin\mathcal{R}
\end{cases}
\end{equation}
Note that $f\left(\bm{Z}\right)$ is a compactly supported function
and is zero outside of the domain $\mathcal{D}$. Moreover, it can
be seen that $f\left(\bm{Z}\right)$ is continuous everywhere. However,
by simple calculation we can find the gradient of $f$ and passing
to the limit $h\to0$, we find that $\lim_{h\to0}\nabla_{_{\bm{Z}}}f$
suffers a finite jump as
\begin{equation}
\lim_{h\to0}\nabla_{_{\bm{Z}}}f=\begin{cases}
-\theta\bm{L}_{3} & ,\bm{Z}\in\mathcal{R}_{+}\\
\left(1-\theta\right)\bm{L}_{3} & ,\bm{Z}\in\mathcal{R}_{0}\\
-\theta\bm{L}_{3} & ,\bm{Z}\in\mathcal{R}_{-}\\
0 & ,\bm{Z}\notin\mathcal{R}
\end{cases}
\end{equation}
in which we used the fact that passing to the limit $h\to0$ results
in $\xi_{1},\xi_{2}\to0$. Define functions $\bm{\Phi}\left(\bm{Z}\right)$ and $\bm{w}\left(\bm{Z}\right)=\text{ax}\left(\bm{W}\left(\bm{Z}\right)\right)$
in (\ref{compactly supported deformations}) and (\ref{Cayley Representation})
as
\begin{equation}
\bm{\Phi}\left(\bm{Z}\right)=\bm{a}f\left(\bm{Z}\right)\qquad\qquad\text{and}\qquad\qquad\bm{w}\left(\bm{Z}\right)=\bm{b}f\left(\bm{Z}\right)
\end{equation}
in which $\bm{a}$ and $\bm{b}$ are fixed vectors in the current
configuration. Define
\begin{equation}
\mathcal{F}=\frac{1}{V\left(\mathcal{R}\right)}\int_{\mathcal{\mathcal{R}}}\left[W\left(\bm{F}_{0e}+\nabla_{_{\bm{Z}}}\bm{\Phi},\bm{d}_{0e},\nabla\bm{d}_{0e}+\frac{\partial\bm{w}}{\partial Z_{A}}\times\bm{d}_{0e}\otimes\bm{E}_{A}\right)-W\left(\bm{F}_{0e},\bm{d}_{0e},\nabla\bm{d}_{0e}\right)\right]dV_{Z}
\end{equation}
in which $V\left(\mathcal{R}\right)$ is the whole volume of the domain
$\mathcal{R}$ and according to (\ref{Quasiconvexity with axial vector}),
we conclude that $\mathcal{F}\geq0$. Now we find rank--one convexity
condition by calculating the following limit
\begin{equation}
\lim_{h\to0}\mathcal{F}\geq0
\end{equation}
 Note that $V\left(\mathcal{R}_{+}\right)=V\left(\mathcal{R}_{-}\right)=\left(1-\theta\right)V\left(\mathcal{R}\right)/2$
and $V\left(\mathcal{R}_{0}\right)=\theta V\left(\mathcal{R}\right)$
as a result, we can write $\mathcal{F}$ as
\begin{align*}
\mathcal{F} & =\frac{\left(1-\theta\right)}{2V\left(\mathcal{R}_{+}\right)}\int_{\mathcal{\mathcal{R}}_{+}}\left[W\left(\bm{F}_{0e}+\nabla_{_{\bm{Z}}}\bm{\Phi},\bm{d}_{0e},\nabla\bm{d}_{0e}+\frac{\partial\bm{w}}{\partial\xi_{A}}\times\bm{d}_{0e}\otimes\bm{E}_{A}\right)-W\left(\bm{F}_{0e},\bm{d}_{0e},\nabla\bm{d}_{0e}\right)\right]dV_{Z}\\
 & +\frac{\theta}{V\left(\mathcal{R}_{0}\right)}\int_{\mathcal{\mathcal{R}}_{0}}\left[W\left(\bm{F}_{0e}+\nabla_{_{\bm{Z}}}\bm{\Phi},\bm{d}_{0e},\nabla\bm{d}_{0e}+\frac{\partial\bm{w}}{\partial Z_{A}}\times\bm{d}_{0e}\otimes\bm{E}_{A}\right)-W\left(\bm{F}_{0e},\bm{d}_{0e},\nabla\bm{d}_{0e}\right)\right]dV_{Z}\\
 & +\frac{\left(1-\theta\right)}{2V\left(\mathcal{R}_{-}\right)}\int_{\mathcal{\mathcal{R}}_{-}}\left[W\left(\bm{F}_{0e}+\nabla_{_{\bm{Z}}}\bm{\Phi},\bm{d}_{0e},\nabla\bm{d}_{0e}+\frac{\partial\bm{w}}{\partial Z_{A}}\times\bm{d}_{0e}\otimes\bm{E}_{A}\right)-W\left(\bm{F}_{0e},\bm{d}_{0e},\nabla\bm{d}_{0e}\right)\right]dV_{Z}.
\end{align*}

Passing to the limit $h\to0$ and invoking the mean value theorem
of integrals, we find
\begin{align*}
\lim_{h\to0}\mathcal{F} & =\left(1-\theta\right)W\left(\bm{F}_{0e}-\theta\bm{a}\otimes\bm{\mathcal{S}}^{T}\bm{L}_{3},\bm{d}_{0e},\nabla\bm{d}_{0e}-\theta\bm{b}\times\bm{d}_{0e}\otimes\bm{\mathcal{S}}^{T}\bm{L}_{3}\right)\\
 & +\theta W\left(\bm{F}_{0e}+\left(1-\theta\right)\bm{a}\otimes\bm{\mathcal{S}}^{T}\bm{L}_{3},\bm{d}_{0e},\nabla\bm{d}_{0e}+\left(1-\theta\right)\bm{b}\times\bm{d}_{0e}\otimes\bm{\mathcal{S}}^{T}\bm{L}_{3}\right)\\
 & -W\left(\bm{F}_{0e},\bm{d}_{0e},\nabla\bm{d}_{0e}\right)\geq0
\end{align*}
where we used the relation $\bm{E}_{3}=\bm{\mathcal{S}}^{T}\bm{L}_{3}$.
By expanding the following equation in terms of $\theta$, we find
\begin{align}
W\left(\bm{F}_{0e}-\theta\bm{a}\otimes\bm{\mathcal{S}}^{T}\bm{L}_{3},\bm{d}_{0e},\nabla\bm{d}_{0e}-\theta\bm{b}\times\bm{d}_{0e}\otimes\bm{\mathcal{S}}^{T}\bm{L}_{3}\right)  &=W\left(\bm{F}_{0e},\bm{d}_{0e},\nabla\bm{d}_{0e}\right)\nonumber \\
 & -\theta\bm{P}_{0e}\cdot\bm{a}\otimes\bm{\mathcal{S}}^{T}\bm{L}_{3}-\theta\bm{M}_{0e}\cdot\bm{b}\times\bm{d}_{0e}\otimes\bm{\mathcal{S}}^{T}\bm{L}_{3} +o\left(\theta^{2}\right)
\end{align}
in which $\bm{P}_{0e}=\bm{P}\left(\bm{F}_{0e},\bm{d}_{0e},\nabla\bm{d}_{0e}\right)$
and $\bm{M}_{0e}=\bm{M}\left(\bm{F}_{0e},\bm{d}_{0e},\nabla\bm{d}_{0e}\right)$.
By substituting above in $\lim_{h\to0}\mathcal{F}$, we find and dividing
the result by $\theta$ and passing to the limit $\theta\to0$, we
find

\begin{align}
W\left(\bm{F}_{0e}+\bm{a}\otimes\bm{\mathcal{S}}^{T}\bm{L}_{3},\bm{d}_{0e},\nabla\bm{d}_{0e}+\bm{b}\times\bm{d}_{0e}\otimes\bm{\mathcal{S}}^{T}\bm{L}_{3}\right)\nonumber \\
-W\left(\bm{F}_{0e},\bm{d}_{0e},\nabla\bm{d}_{0e}\right) & \geq\bm{P}_{0e}\cdot\bm{a}\otimes\bm{\mathcal{S}}^{T}\bm{L}_{3}+\bm{M}_{0e}\cdot\bm{b}\times\bm{d}_{0e}\otimes\bm{\mathcal{S}}^{T}\bm{L}_{3}
\end{align}

Furthermore, given that $\bm{L}_{3}$ and $\bm{\mathcal{S}}$ are
arbitrary, we conclude that $\bm{\mathcal{S}}^{T}\bm{L}_{3}$ is also
an arbitrary vector. Therefore, we can use any other arbitrary vector
$\bm{T}$ and rewrite the above relation as

\begin{equation}
W\left(\bm{F}_{0e}+\bm{a}\otimes\bm{T},\bm{d}_{0e},\nabla\bm{d}_{0e}+\bm{b}\times\bm{d}_{0e}\otimes\bm{T}\right)-W\left(\bm{F}_{0e},\bm{d}_{0e},\nabla\bm{d}_{0e}\right)\geq\bm{P}_{0e}\cdot\bm{a}\otimes\bm{T}+\bm{M}_{0e}\cdot\bm{b}\times\bm{d}_{0e}\otimes\bm{T}\label{Rank1 convexity}
\end{equation}

Relation (\ref{Rank1 convexity}) is called a rank--one convexity condition
that must be satisfied at each material point for any arbitrary vectors
$\bm{a}$, $\bm{b}$ in the current configuration and $\bm{T}$ in
the reference configuration by $\bm{\chi}_{e}$ and $\bm{d}_{e}$.
Next, we find Legendre--Hadamard inequality. 

\subsubsection{Legendre--Hadamard inequalities} \label{Legendre-Hadamard inequalities}

Consider the following vectors used in (\ref{Rank1 convexity})
\begin{equation}
\bm{a}=\epsilon\bm{s}\qquad\qquad\text{and}\qquad\qquad\bm{b}=\epsilon\bm{q}
\end{equation}
where $\epsilon \ll 1$ is a fix scalar, and expand the following energy function $W\left(\bm{F}_{0e}+\epsilon\bm{s}\otimes\bm{T},\bm{d}_{0e},\nabla\bm{d}_{0e}+\epsilon\bm{q}\times\bm{d}_{0e}\otimes\bm{T}\right)$
in terms of $\epsilon$ as
\begin{align}
& W\left(\bm{F}_{0e}+\epsilon\bm{s}\otimes\bm{T},\bm{d}_{0e},\nabla\bm{d}_{0e}+\epsilon\bm{q}\times\bm{d}_{0e}\otimes\bm{T}\right)  -W\left(\bm{F}_{0e},\bm{d}_{0e},\nabla\bm{d}_{0e}\right)= \nonumber \\
 +&\epsilon\left(\bm{P}_{0e}\cdot\bm{s}\otimes\bm{T}+\bm{M}_{0e}\cdot\bm{q}\times\bm{d}_{0e}\otimes\bm{T}\right) \nonumber \\
  +&\frac{\epsilon^{2}}{2}\left(\bm{\mathscr{M}}\left[\bm{s}\otimes\bm{T}\right]\cdot\bm{s}\otimes\bm{T}+2\bm{\mathscr{N}}\left[\bm{s}\otimes\bm{T}\right]\cdot\bm{q}\times\bm{d}_{0e}\otimes\bm{T}+\bm{\mathscr{L}}\left[\bm{q}\times\bm{d}_{0e}\otimes\bm{T}\right]\cdot\bm{q}\times\bm{d}_{0e}\otimes\bm{T}\right)&\nonumber \\
 +&o\left(\epsilon^{2}\right).\label{expansion}
\end{align}
in which
\begin{align}
\bm{\mathscr{M}} & =\left(\frac{\partial^{2}W}{\partial\bm{F}^{2}}\right)_{\mid\bm{d}=\bm{d}_{0e},\bm{\chi}=\bm{\chi}_{0e}}\\
\bm{\mathscr{N}} & =\left(\frac{\partial^{2}W}{\partial\bm{F}\partial\nabla\bm{d}}\right)_{\mid\bm{d}=\bm{d}_{0e},\bm{\chi}=\bm{\chi}_{0e}}=\left(\frac{\partial^{2}W}{\partial\nabla\bm{d}\partial\bm{F}}\right)_{\mid\bm{d}=\bm{d}_{0e},\bm{\chi}=\bm{\chi}_{0e}}\\
\bm{\mathscr{L}} & =\left(\frac{\partial^{2}W}{\partial\left(\nabla\bm{d}\right)^{2}}\right)_{\mid\bm{d}=\bm{d}_{0e},\bm{\chi}=\bm{\chi}_{0e}}
\end{align}
By substituting (\ref{expansion}) into rank--one convexity condition
(\ref{Rank1 convexity}), and dividing the result by $\epsilon^{2}$ and
passing to the limit $\epsilon\to0$ we find
\begin{equation}
\bm{\mathscr{M}}\left[\bm{s}\otimes\bm{T}\right]\cdot\bm{s}\otimes\bm{T}+2\bm{\mathscr{N}}\left[\bm{s}\otimes\bm{T}\right]\cdot\bm{q}\times\bm{d}_{0e}\otimes\bm{T}+\bm{\mathscr{L}}\left[\bm{q}\times\bm{d}_{0e}\otimes\bm{T}\right]\cdot\bm{q}\times\bm{d}_{0e}\otimes\bm{T}\geq0\label{LH Condition}
\end{equation}

Relation (\ref{LH Condition}) is called the Legendre--Hadamard inequality
that must be satisfied for any arbitrary vectors $\bm{s}$, $\bm{q}$
and $\bm{T}$ by the solutions of (\ref{Balance of Linear Momentum})
and (\ref{Balance of Director Momentum}). The strong form of the Legendre--Hadamard inequality is called the ellipticity condition. Violation of the ellipticity condition, also known as loss of ellipticity, is associated with material instabilities.

\section{Numerical simulations} \label{Numerical simulations}
In this section, we present numerical solutions by using Finite Element Analysis (FEA) and two multilayer perceptron networks.
\subsection{Constitutive relation} \label{Constitutive relation}

In this section, we will use the following elastic energy as a representative
for a generic elastic energy to capture the mechanical behavior of
directed media. This energy is inspired by those energies used for nematic elastomers introduced in \cite{zhou2025modified, wei2024exceptional}. 
The simulations consider pulling a two--dimensional object embedded in two--dimensional space (to be able to compare the solutions we get from both the neural network and FEA), as a result, we use the following form of energy
\begin{equation}
W=\frac{\mu}{2}(\text{tr}(\bm{L}_{0}\bm{F}^{T}\bm{L}^{-1}\bm{F})-2-2\ln J)+\frac{\alpha\gamma^{2}}{2}\nabla\bm{d}\cdot\nabla\bm{d} \label{nematic elastic energy}
\end{equation}
 where $\mu$ and $\alpha$ are the shear modulus and Frank elastic
modulus, respectively, $\gamma$ is the size effect, $J=\det\bm{F}$,
and $\bm{L}$ and $\bm{L}_{0}$ are defined as 
\begin{equation}
\bm{L}=\bm{I}+(r-1)\bm{d}\otimes\bm{d}\qquad\qquad\text{and}\qquad\qquad\bm{L}_{0}=\bm{I}+(r-1)\bm{D}\otimes\bm{D}
\end{equation}
in which $\bm{D}$ is the initial orientation of the director field
in the reference configuration, $r = \frac{\ell_{\parallel}}{\ell_{\perp}}$ is the shape anisotropy of the network in which $\ell_{\perp}$ and $\ell_{\parallel}$
are material parameters that
correspond to the effective step lengths perpendicular and parallel
to the director, respectively \cite{zhou2025modified, wei2024exceptional}.

In all simulations reported in the rest of this work, the normalized material parameters are taken as
\begin{equation}
\mu=1,\qquad r=2,\qquad\frac{\gamma^{2}\alpha}{2}=5\times10^{-4},
\end{equation}

Before we perform numerical analysis,
we show that this elastic energy satisfies (\ref{Objectivity
symmetric}) and (\ref{LH Condition})
\textit{a priori}. Therefore, the loss function is
frame invariant and the solutions that the network produces satisfy these necessary conditions, specifically, Legendre--Hadamard inequalities.
By using (\ref{stresses}), we find
\begin{align}\bm{P} & =\mu\left(\bm{L}^{-1}\bm{F}\bm{L}_{0}-\bm{F}^{-T}\right)\\
\bm{N} & =\mu\left(\frac{1}{r}-1\right)\left(\bm{F}\bm{L}_{0}\bm{F}^{T}\right)\bm{d}\\
\bm{M} & =\alpha\gamma^{2}\nabla\bm{d}
\end{align}
then by substituting these stresses and moments into (\ref{Objectivity
symmetric}), we can see that
\begin{equation}
\bm{P}\bm{F}^{T}+\bm{N}\otimes\bm{d}+\bm{M}\left(\nabla\bm{d}\right)^{T}=\mu\left(\bm{L}^{-1}\bm{F}\bm{L}_{0}\bm{F}^{T}+\bm{F}\bm{L}_{0}\bm{F}^{T}\bm{L}^{-1}-\bm{F}\bm{L}_{0}\bm{F}^{T}-\bm{I}\right)+\alpha\gamma^{2}\nabla\bm{d}\left(\nabla\bm{d}\right)^{T}\in\text{Sym}
\end{equation}

Furthermore, one can show that 
\begin{equation}
\bm{\mathscr{M}}\left[\bm{s}\otimes\bm{T}\right]\cdot\bm{s}\otimes\bm{T}=\mu\left(\left(\bm{s}\cdot\bm{L}^{-1}\bm{s}\right)\left(\bm{T}\cdot\bm{L}_{0}\bm{T}\right)+2J^{-1}\left(\bm{s}\cdot\bm{F}^{-T}\bm{T}\right)^{2}\right)
\end{equation}
in which by picking $\bm{s}$ and $\bm{T}$ to be unit vectors we
find
\begin{align}
\left(\bm{s}\cdot\bm{L}^{-1}\bm{s}\right)\left(\bm{T}\cdot\bm{L}_{0}\bm{T}\right)&=\frac{1}{4r}\left(r+1-\left(r- 1\right)\cos2\lambda\right)\left(r+1 + \left(r-1\right)\cos2\lambda_{0}\right)
\end{align}
where $\cos\lambda=\bm{s}\cdot\bm{d}$ and $\cos\lambda_{0}=\bm{T}\cdot\bm{D}$.
In this example, we have with $r = \frac{\ell_{\parallel}}{\ell_{\perp}}=2$
and as a result, we find
\begin{equation}
\left(\bm{s}\cdot\bm{L}^{-1}\bm{s}\right)\left(\bm{T}\cdot\bm{L}_{0}\bm{T}\right)=\frac{1}{8}\left(3-\cos2\lambda\right)\left(3+\cos2\lambda_{0}\right) > 0
\end{equation}
from which, we can see that the following strong ellipticity condition, which is the strong form of the Legendre--Hadamard inequality
is satisfied
\begin{equation}
\bm{\mathscr{M}}\left[\bm{s}\otimes\bm{T}\right]\cdot\bm{s}\otimes\bm{T}>0. \label{LH1 for the loss function example}
\end{equation}
Therefore, the adopted constitutive model satisfies the strong ellipticity condition, ensuring material stability within the admissible range of deformation states considered. Similarly, it can be seen that
\begin{equation}
\bm{\mathscr{L}}\left[\bm{q}\times\bm{d}_{e0}\otimes\bm{T}\right]\cdot\bm{q}\times\bm{d}_{e0}\otimes\bm{T}>0 \label{LH2 for the loss function example}
\end{equation}
and in consequence, the solutions obtained by the network satisfy
the Legendre--Hadamard inequalities \textit{a priori} and hence, for the chosen constitutive model and parameter set, the ellipticity condition holds within the regime considered for each material point.  It can be seen from (\ref{LH1 for the loss function example}) and (\ref{LH2 for the loss function example}) that the ellipticity conditions are satisfied and as a result the material model is strongly elliptic for the selected parameters.

\subsection{Finite Element Analysis} \label{Finite element formulation}

Finite element simulations are used to compute equilibrium configurations of a directed elastic medium and provide a reference solution for comparison with the neural network--based solver. The simulations are carried out using the open--source finite element framework FEniCSx \cite{Alnaes2015}, which provides an automated environment for the variational formulation and solution of nonlinear partial differential equations. The numerical implementation follows the variational formulation introduced in section (\ref{Balance laws and boundary conditions}) and computes stationary points of the elastic energy defined in (\ref{nematic elastic energy}). The reference domain is taken as the rectangular region

\begin{equation}
\kappa_0 = [0,L] \times [0,W],
\qquad
L = 1.0, \qquad W = 0.2,
\end{equation}
where all dimensions are normalized by $H_0 = 1$ cm. The reference domain is discretized using a structured quadrilateral mesh with $N_x = 25$ elements in the longitudinal direction and $N_y = 5$ elements in the transverse direction. 

The unknown fields are the displacement $\bm{u}$ and a single director field $\bm{d}$. In two dimensions, the director is represented by an angular field
\begin{equation}
\varphi : \kappa_0 \rightarrow \mathbb{R},
\qquad
\bm{d} = (\cos\varphi,\sin\varphi),
\label{director for numerical simulations}
\end{equation}
which enforces the unit length constraint $|\bm{d}|=1$ identically. The unknown fields are approximated using continuous Lagrange finite elements,
\begin{equation}
\bm{u}_h \in [\mathcal P_2]^2,
\qquad
\varphi_h \in \mathcal P_2 .
\end{equation}
Numerical integration of all surface and boundary line integrals is performed using quadrature degree 4. For notational convenience, we denote the four sides of the rectangle by
\begin{equation}
\Gamma_L=\{X=0\},\quad
\Gamma_R=\{X=L\},\quad
\Gamma_B=\{Y=0\},\quad
\Gamma_T=\{Y=W\}.
\end{equation}

Displacement--controlled loading is applied through Dirichlet boundary conditions,
\begin{align}
\bm{u} &= (0,0) && \text{on } \Gamma_L, \\
\bm{u} &= (\Delta u(t), 0) && \text{on } \Gamma_R, \\
\varphi &= \varphi_0 && \text{on } \Gamma_L.
\end{align}
where load parameter $t$ denotes a pseudo--time (increment counter) used to apply the displacement in $N$ equal increments, and $\Delta u(t)$ is
\begin{equation}
\Delta u(t) = \Delta u_{\max}\,\frac{t}{N},
\qquad
\Delta u_{\max} = 0.1\,L,
\qquad
N = 300 .
\end{equation}

The director field is initialized uniformly in the reference configuration as
\begin{equation}
\varphi(X) = \varphi_0
\qquad
\forall X \in \kappa_0,
\end{equation}
which is consistent with the prescribed boundary condition on $\Gamma_L$.

At each load increment, equilibrium is obtained through a staggered minimization of the total energy. First, the displacement field is solved for fixed director field by enforcing
\begin{equation}
\delta \Pi(\bm{u},\varphi;\delta \bm{u})=0,
\end{equation}
using a Newton iteration. Next, the director field is updated for fixed displacement by solving
\begin{equation}
\delta \Pi(\bm{u},\varphi;\delta \varphi)=0.
\end{equation}
This alternating procedure is repeated at each increment of the prescribed displacement.

All nonlinear problems are solved using Newton's method with consistent linearization obtained via automatic differentiation. Convergence is assessed using an incremental criterion with relative tolerance $10^{-6}$. Linearized systems are solved using Krylov subspace methods with standard algebraic preconditioning.

\subsubsection{Mesh Independence Study}
A mesh independence study was conducted for the case where $\varphi_0 = \pi/3$ to ensure that the numerical solution was independent of mesh resolution. The number of discretizations in the y dimension ($n_y$) ranged from 5 to 20. The number of corresponding discretizations in the $x$ dimension was equal to $5\times n_y$. The director orientation response is effectively mesh--converged over the tested range, as the min, mean, and max director orientation angles are unchanged with refinement.  Therefore, conclusions based on these results are mesh--independent across these meshes.

\begin{figure}[h]
    \centering
    \includegraphics[width=0.6\textwidth]{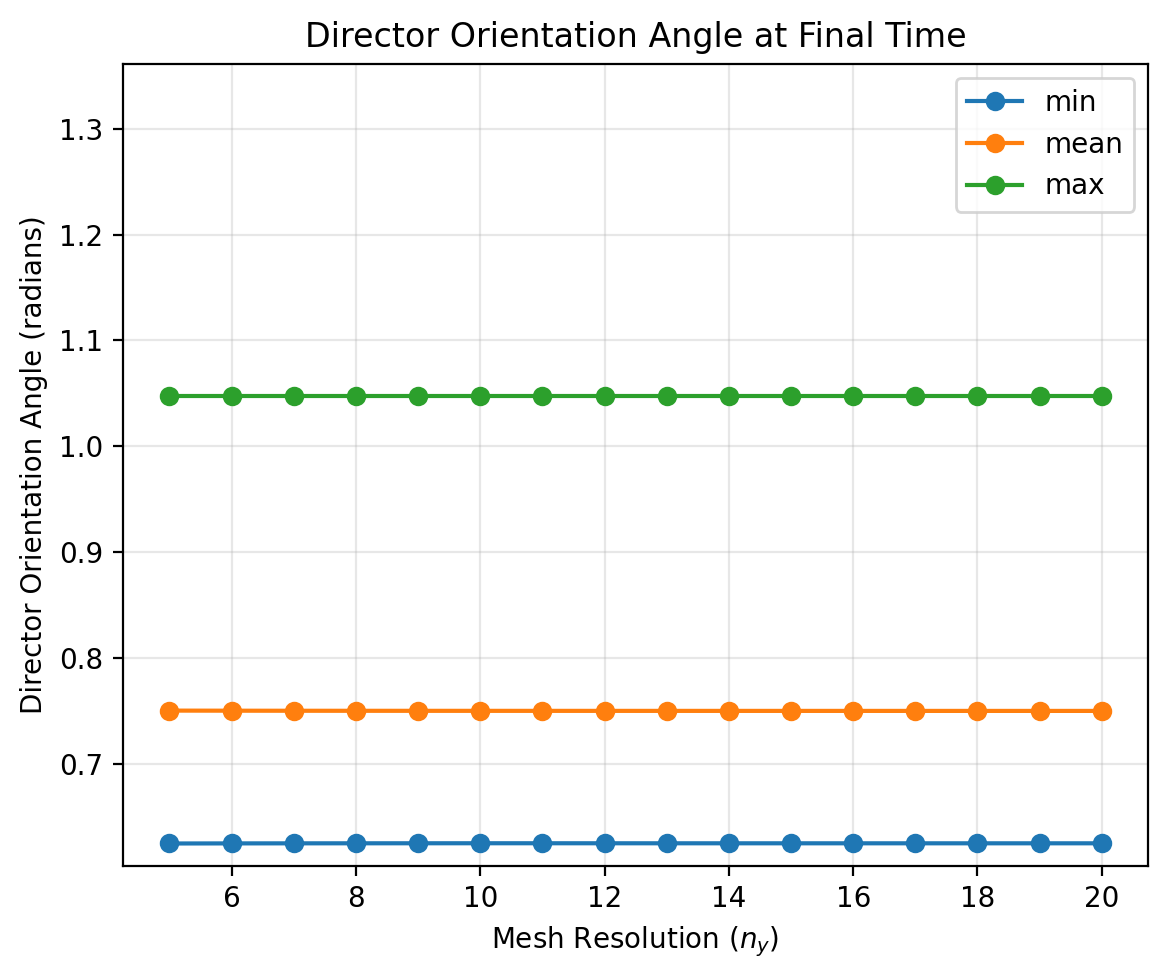}
    \caption{Mesh--independence study showing the minimum, mean, and maximum values of director orientation angle at the final simulation time as a function of mesh resolution ($n_y$).}
    \label{Mesh Independence Study Results}
    \end{figure}

\subsection{Neural Network--Based Solver}

\begin{figure}[h]
    \centering
    \includegraphics[width=0.95\textwidth]{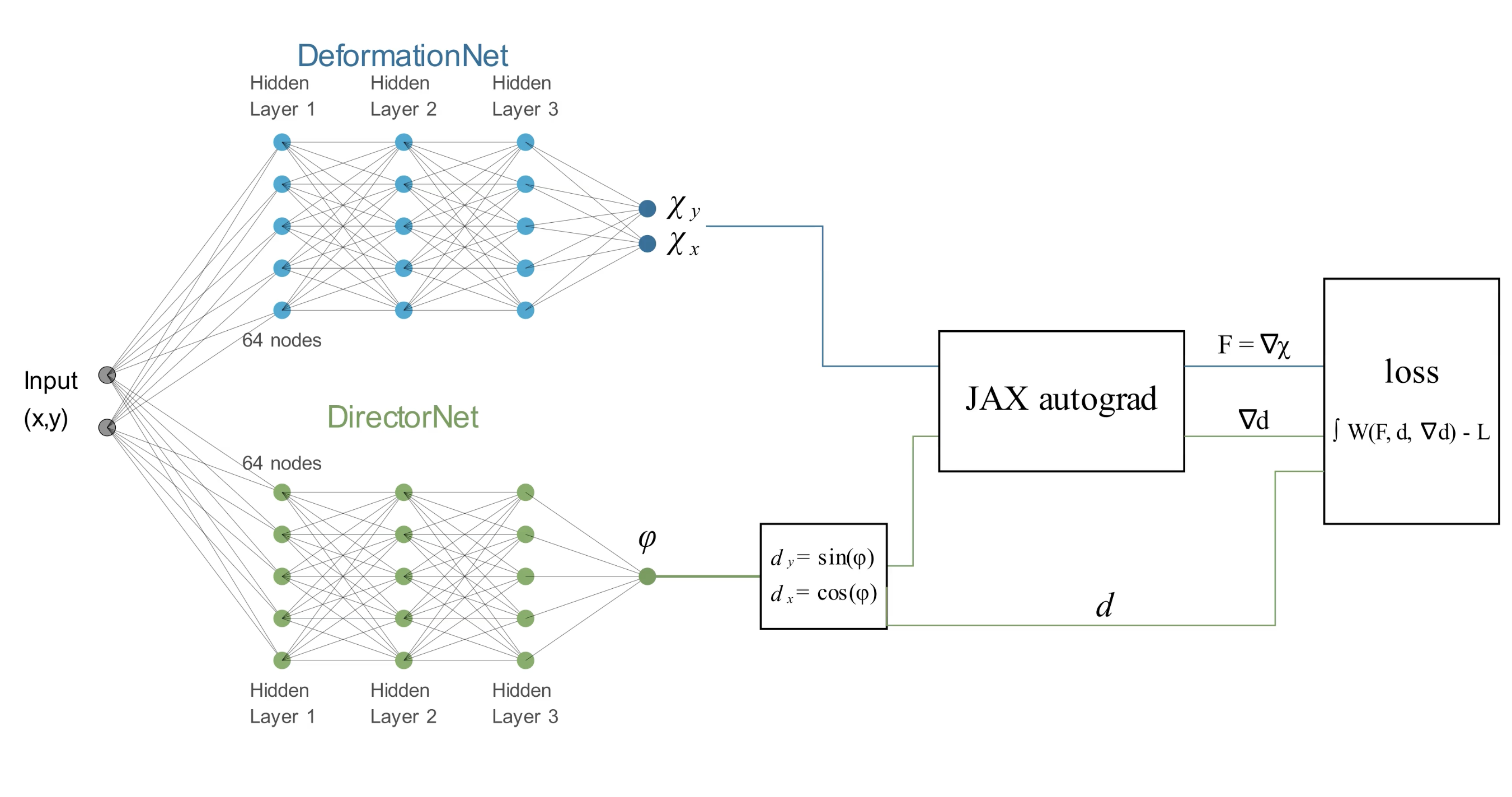}
    \caption{Neural Network Architecture with DeformationNet and DirectorNet}
    \label{NNArchitecture}
\end{figure}

In this section, we present a neural network--based approach to obtain the solution of materials with microstructures under deformation. The kinematical fields in this problem are the deformation map $\bm{\chi}$ and the director field $\bm{d}$. In principle, one could construct a single network that maps material points to both $\bm{\chi}$ and $\bm{d}$ while minimizing the total potential energy. However, because the deformation and director fields are kinematically independent and should be varied independently in the minimization process, we instead employ two separate neural networks. The overall architecture is shown in Fig. (\ref{NNArchitecture}).

In this figure, both networks take a material point in the reference configuration denoted by $\bm{X}=(X,Y)\in\Omega\subset\mathbb{R}^2$ as an input. The \textit{DeformationNet} generates the displacement field ($\bm{u}(\bm{X})$), from which the deformation map $\bm{\chi}=\bm{X}+\bm{u}$ is obtained. The \textit{DirectorNet} generates a scalar director angle ($\varphi(\bm{X})$), from which the director $\bm{d}$ is constructed using (\ref{director for numerical simulations}). By this construction, the constraint (\ref{constraint}) is automatically satisfied.

\noindent \textbf{Deformation Network.} The raw displacement output $\tilde{\bm{u}}_\theta(\bm{X})\in\mathbb{R}^2$ is parameterized by a fully--connected neural network with three hidden layers of width 64 and $\tanh$ activation functions
\begin{align}
\bm{h}^{u}_{1} &= \tanh\!\left(\bm{W}^{u}_{1}\bm{X}+\bm{b}^{u}_{1}\right), \\
\bm{h}^{u}_{2} &= \tanh\!\left(\bm{W}^{u}_{2}\bm{h}^{u}_{1}+\bm{b}^{u}_{2}\right), \\
\bm{h}^{u}_{3} &= \tanh\!\left(\bm{W}^{u}_{3}\bm{h}^{u}_{2}+\bm{b}^{u}_{3}\right), \\
\tilde{\bm{\chi}}_\theta(\bm{X}) &= \bm{W}^{u}_{4}\bm{h}^{u}_{3}+\bm{b}^{u}_{4}.
\end{align}

The trainable parameter set is $\theta=\{\bm{W}^{u}_{\ell},\bm{b}^{u}_{\ell}\}_{\ell=1}^{4}$.

\noindent \textbf{Director Network.}
The \textit{DirectorNet} produces a scalar angle field $\tilde{\varphi}_\eta(\bm{X})\in\mathbb{R}$
\begin{align}
\bm{h}^{\theta}_{1} &= \tanh\!\left(\bm{W}^{\theta}_{1}\bm{X}+\bm{b}^{\theta}_{1}\right), \\
\bm{h}^{\theta}_{2} &= \tanh\!\left(\bm{W}^{\theta}_{2}\bm{h}^{\theta}_{1}+\bm{b}^{\theta}_{2}\right), \\
\bm{h}^{\theta}_{3} &= \tanh\!\left(\bm{W}^{\theta}_{3}\bm{h}^{\theta}_{2}+\bm{b}^{\theta}_{3}\right), \\
\tilde{\varphi}_\eta(\bm{X}) &= \bm{W}^{\theta}_{4}\bm{h}^{\theta}_{3}+\bm{b}^{\theta}_{4},
\end{align}
with trainable parameters $\eta=\{\bm{W}^{\theta}_{\ell},\bm{b}^{\theta}_{\ell}\}_{\ell=1}^{4}$. The director field $\bm{d}(\bm{X})$ is then constructed as
\begin{equation}
\bm{d}(\bm{X})=
\big(\cos\varphi(\bm{X}),\,\sin\varphi(\bm{X})\big),
\end{equation}
which corresponds to (\ref{director for numerical simulations}) and satisfies the unit length constraint (\ref{constraint}). 

To enforce boundary conditions without adding penalty terms to the loss function, we employ ansatz functions. The physical displacement field is defined as
\begin{equation}
\bm{u}(\bm{X})
=
(\frac{X}{L}\,\Delta L,
0)+X(L-X)\,\tilde{\bm{u}}_\theta(\bm{X}),
\end{equation}
which satisfies the prescribed displacement boundary conditions at $X=0$ and $X=L$. The director angle is constructed as
\begin{equation}
\varphi(\bm{X})
=
\varphi_0 + X\,\tilde{\varphi}_\eta(\bm{X}),
\end{equation}
so that the initial orientation $\varphi_0$ is enforced at $X=0$.

The loss function corresponds to the total potential energy defined in (\ref{total potential energy}), in which the elastic contribution is given in (\ref{nematic elastic energy}). Because the boundary conditions are enforced through the ansatz \cite{herrmann2025deep}, they do not appear explicitly as penalty terms in the loss function.

\subsection{Verification of NN with FEA}

To assess the accuracy of the neural network--based solver, we compare its final equilibrium solutions with those obtained from FEA for three different initial director orientations: $\varphi_0 = \pi / 3$, $\pi / 4$, and $\pi / 6$. In all cases, figures (\ref{pi3} - \ref{pi6}) display the final deformed configuration, including normalized displacement magnitude ($|\bm{u}|^2 = u_x^2 + u_y^2$), normalized displacement components ($u_x$ and $u_y$), and the director orientation $\varphi$  at equilibrium. The parameter $\varphi_0$ refers to the uniform initial director orientation prescribed in the reference configuration prior to loading which remains fixed on the Dirichlet boundary condition ($X=0, Y\in (0, W)$).

\subsubsection{Case 1: $\varphi_0 = \pi / 3$}
\begin{figure}[h]
    \centering

    \begin{subfigure}{\textwidth}
        \centering
        \includegraphics[width=0.95\textwidth]{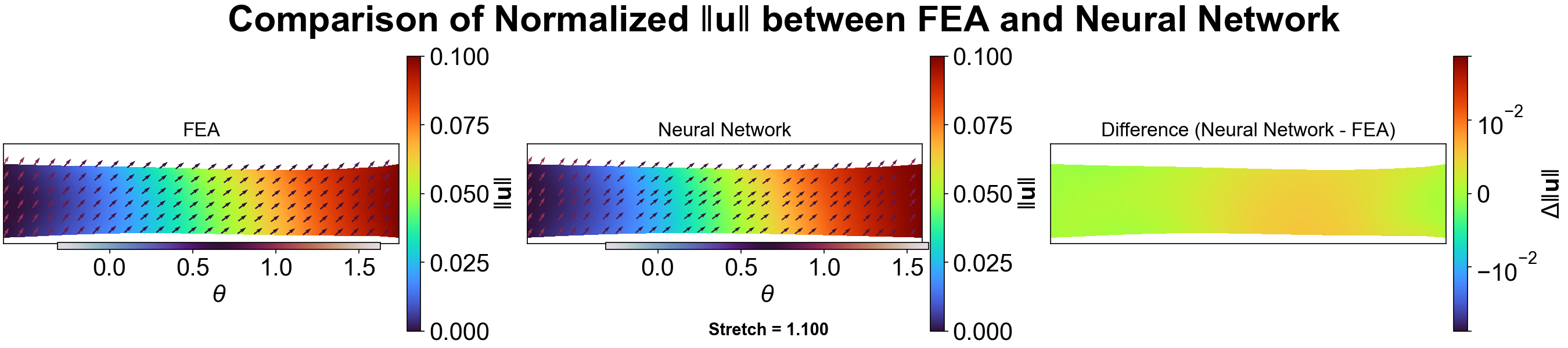}
        \caption{}
    \end{subfigure}\par\vspace{0.4em}

    \begin{subfigure}{\textwidth}
        \centering
        \includegraphics[width=0.95\textwidth]{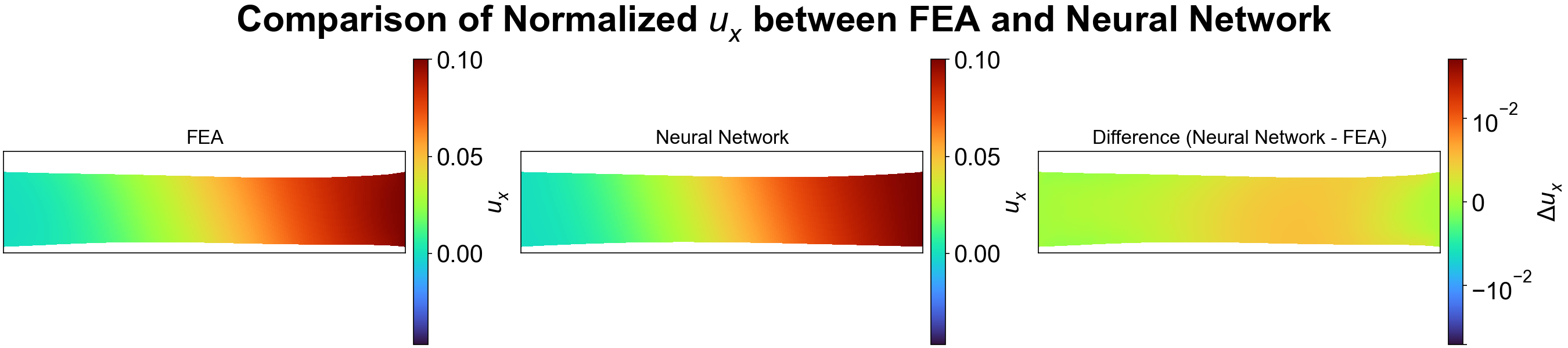}
        \caption{}
    \end{subfigure}\par\vspace{0.4em}

    \begin{subfigure}{\textwidth}
        \centering
        \includegraphics[width=0.95\textwidth]{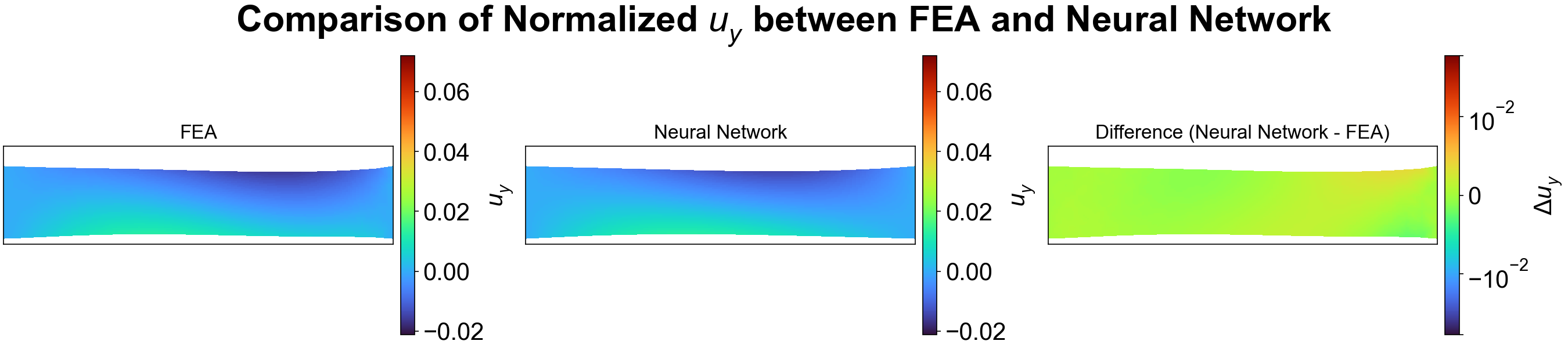}
        \caption{}
    \end{subfigure}\par\vspace{0.4em}

    \begin{subfigure}{\textwidth}
        \centering
        \includegraphics[width=0.95\textwidth]{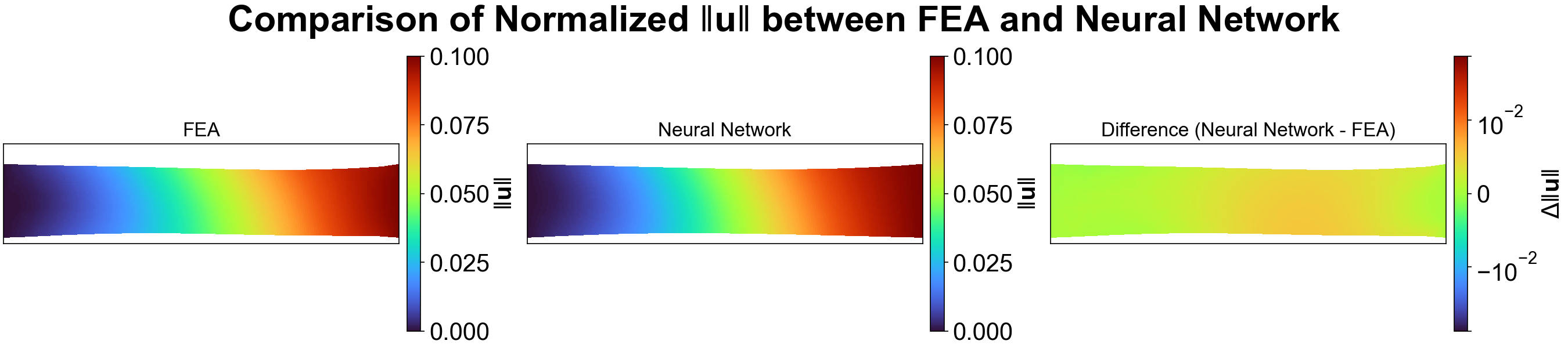}
        \caption{}
    \end{subfigure}\par\vspace{0.4em}

    \begin{subfigure}{\textwidth}
        \centering
        \includegraphics[width=0.95\textwidth]{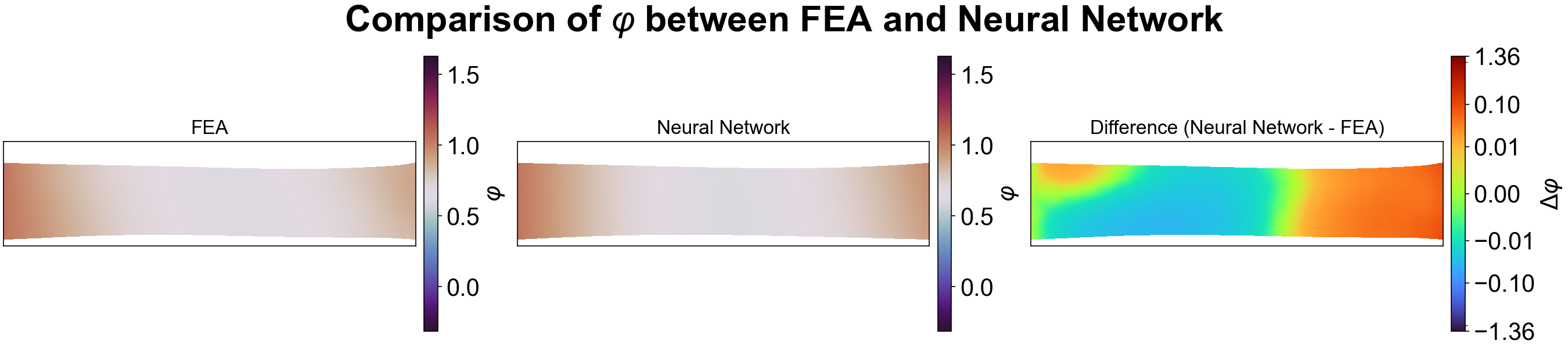}
        \caption{}
    \end{subfigure}

    \caption{Comparison between solutions obtained from FEA and NN for the case with $\varphi_0 = \pi/3$. From the right to the left, results from FEA, results from NN, and the difference between the solutions. (a) Magnitude of deformation with director fields, (b) Magnitude of deformation, (c) Magnitude of deformation in $x-$direction (d) Magnitude of deformation in $y-$direction, and (e) Orientation of the director field.}
    \label{pi3}
\end{figure}

For the case where the director is initially oriented at $\varphi_0 = \pi / 3$ (shown in figure (\ref{pi3})), relative to the x-axis, the imposed right boundary displacement produces a predominant $u_x$ stretch, with secondary transverse effects arising from large deformation and the coupling between deformation and director fields. At equilibrium, the FEA solution shows a smooth x--displacement gradient across the specimen, with $u_x$ dominating the response and $u_y$ remaining smaller but non--negligible.

The neural network reproduces the overall deformation and director fields. The normalized displacement magnitude and spatial gradient in $u_x$ are similar to the FEA solution, with the largest discrepancies appearing in regions where the $X-$gradient is steepest. This is expected, since $u_x$ carries the primary mechanical response and therefore accumulates the largest absolute errors. The transverse displacement field $u_y$ exhibit closer agreement.

Importantly, although the neural network begins training with randomly initialized values, the final predicted $\varphi$ field closely matches the FEA equilibrium configuration. The director reorients smoothly under stretch, and the network captures both the magnitude and spatial variation of this reorientation. The difference plots confirm that deviations are small and spatially smooth, with no spurious oscillations or localized instabilities.

\subsubsection{Case 2: $\varphi_0 = \pi / 4$}

\begin{figure}[h]
    \centering

    \begin{subfigure}{\textwidth}
        \centering
        \includegraphics[width=0.95\textwidth]{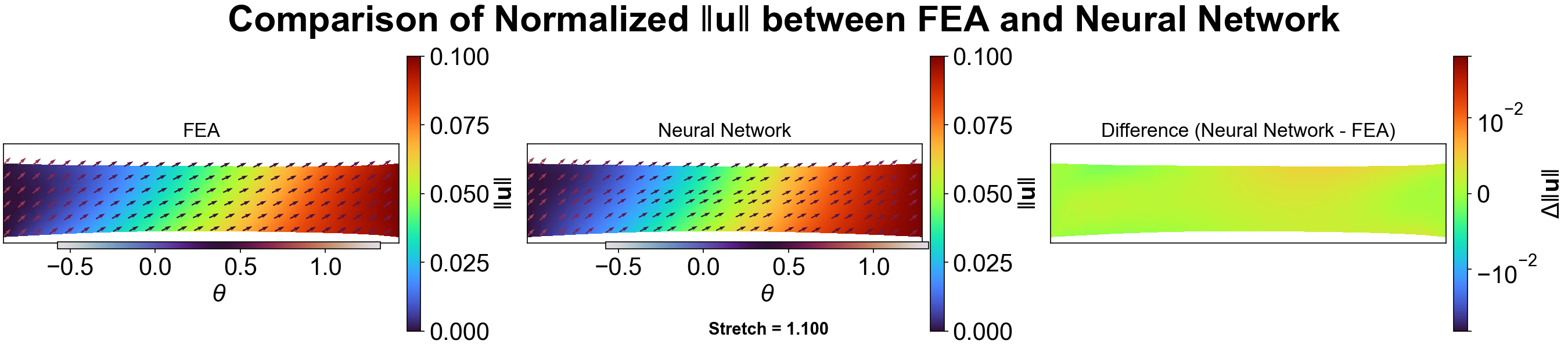}
        \caption{}
    \end{subfigure}\par\vspace{0.4em}

    \begin{subfigure}{\textwidth}
        \centering
        \includegraphics[width=0.95\textwidth]{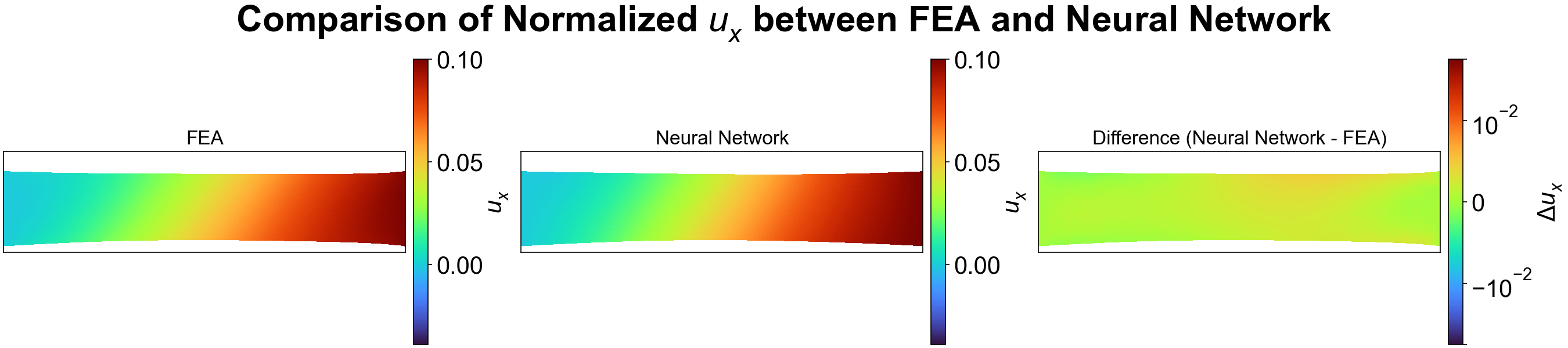}
        \caption{}
    \end{subfigure}\par\vspace{0.4em}

    \begin{subfigure}{\textwidth}
        \centering
        \includegraphics[width=0.95\textwidth]{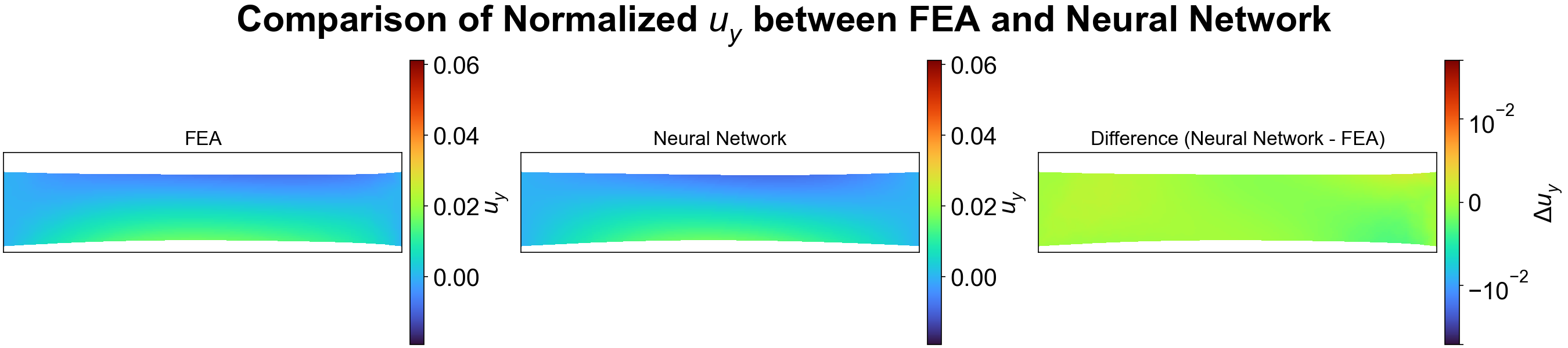}
        \caption{}
    \end{subfigure}\par\vspace{0.4em}

    \begin{subfigure}{\textwidth}
        \centering
        \includegraphics[width=0.95\textwidth]{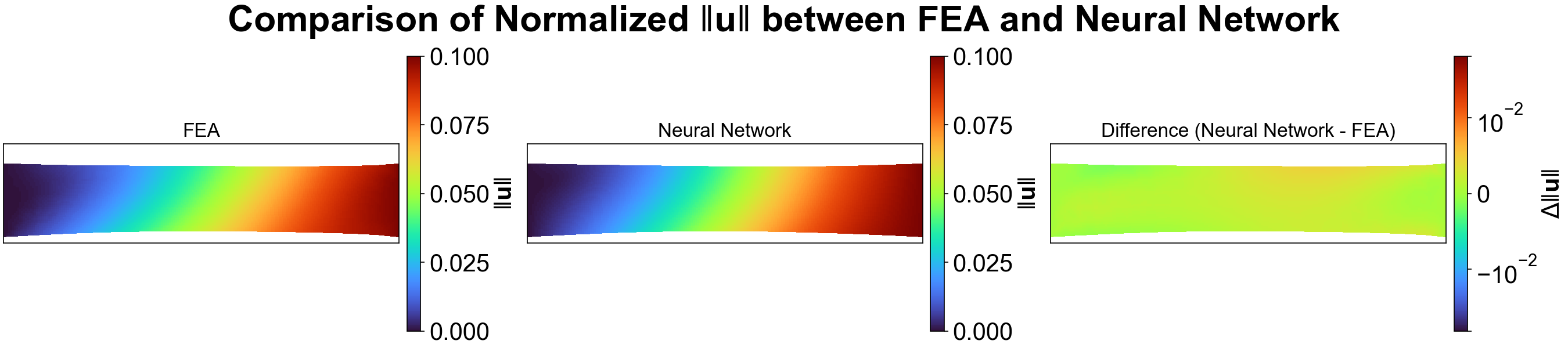}
        \caption{}
    \end{subfigure}\par\vspace{0.4em}

    \begin{subfigure}{\textwidth}
        \centering
        \includegraphics[width=0.95\textwidth]{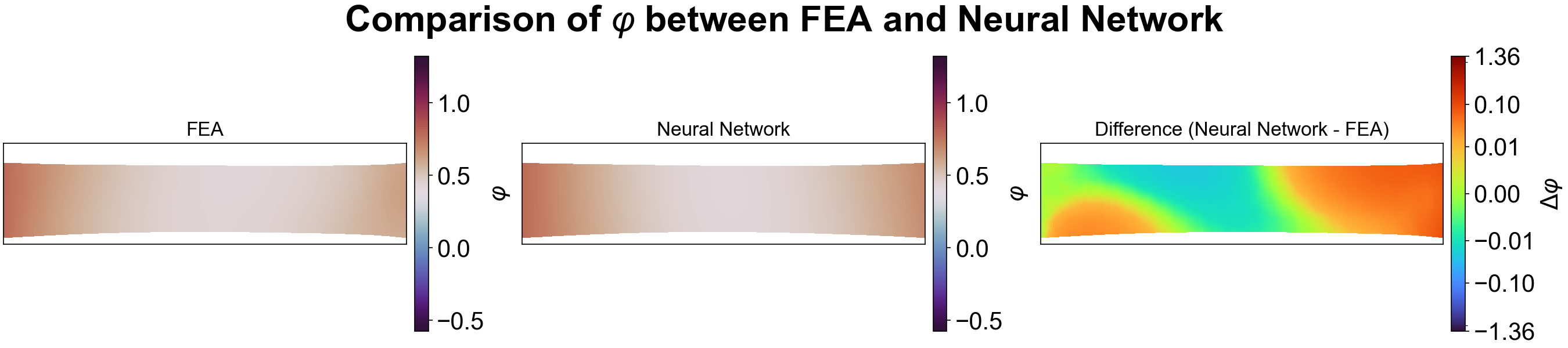}
        \caption{}
    \end{subfigure}

    \caption{Comparison between solutions obtained from FEA and NN for the case with $\varphi_0 = \pi/4$. From the right to the left, results from FEA, results from NN, and the difference between the solutions. (a) Magnitude of deformation with director fields, (b) Magnitude of deformation, (c) Magnitude of deformation in $x-$direction (d) Magnitude of deformation in $y-$direction, and (e) Orientation of the director field.}
    \label{pi4}
\end{figure}

When the director is initially oriented at  $\varphi_0 = \pi / 4$ (shown in figure (\ref{pi4})), it lies midway between the $X$ and $Y$ axes. This configuration results in a more symmetric coupling between deformation and director evolution. In the final equilibrium state, both FEA and neural network solutions show a smooth $u_x$ with modest curvature in $Y$ direction.

In this case, agreement between the two methods is particularly strong. The $u_x$ and $u_y$  fields are nearly indistinguishable at the plotted resolution, and the difference fields remain uniformly small throughout the domain. The final director orientation predicted by the neural network aligns closely with the FEA solution, indicating that the network correctly captures the energetically preferred reorientation under loading. Because the initial orientation is neither strongly aligned nor strongly oblique to the stretch direction, this case provides a balanced test of the coupled kinematics.

\subsubsection{Case 3: $\varphi_0 = \pi / 6$}

\begin{figure}[h]
    \centering

    \begin{subfigure}{\textwidth}
        \centering
        \includegraphics[width=0.95\textwidth]{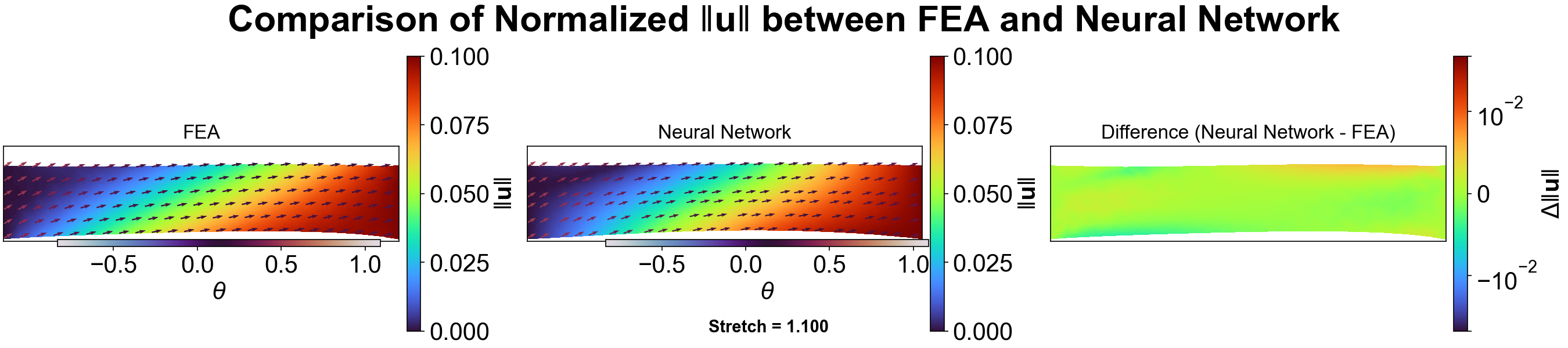}
        \caption{}
    \end{subfigure}\par\vspace{0.4em}

    \begin{subfigure}{\textwidth}
        \centering
        \includegraphics[width=0.95\textwidth]{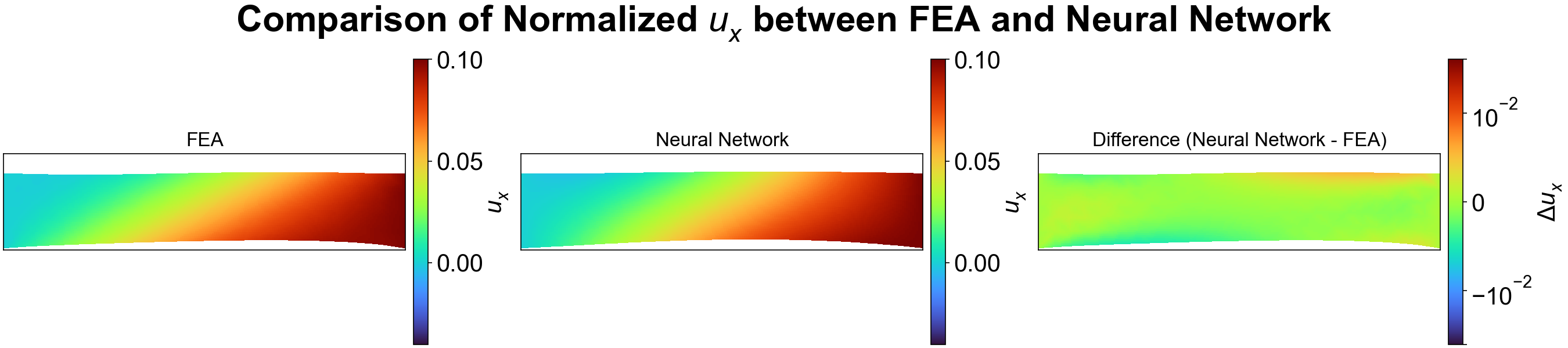}
        \caption{}
    \end{subfigure}\par\vspace{0.4em}

    \begin{subfigure}{\textwidth}
        \centering
        \includegraphics[width=0.95\textwidth]{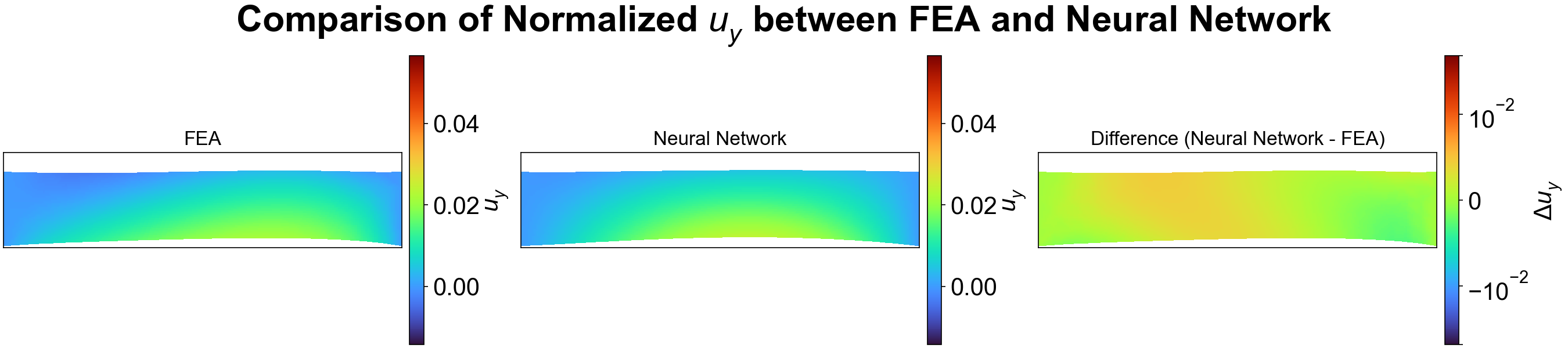}
        \caption{}
    \end{subfigure}\par\vspace{0.4em}

    \begin{subfigure}{\textwidth}
        \centering
        \includegraphics[width=0.95\textwidth]{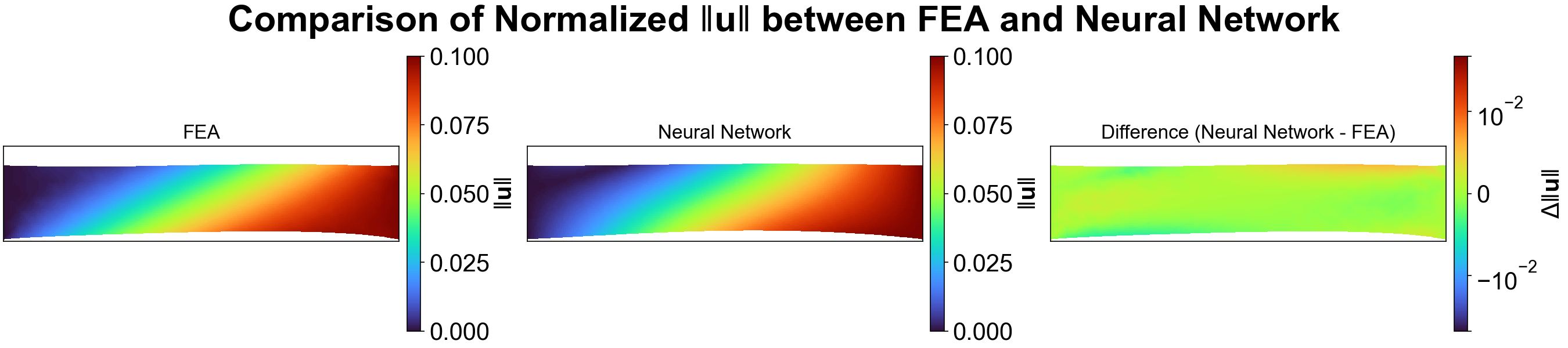}
        \caption{}
    \end{subfigure}\par\vspace{0.4em}

    \begin{subfigure}{\textwidth}
        \centering
        \includegraphics[width=0.95\textwidth]{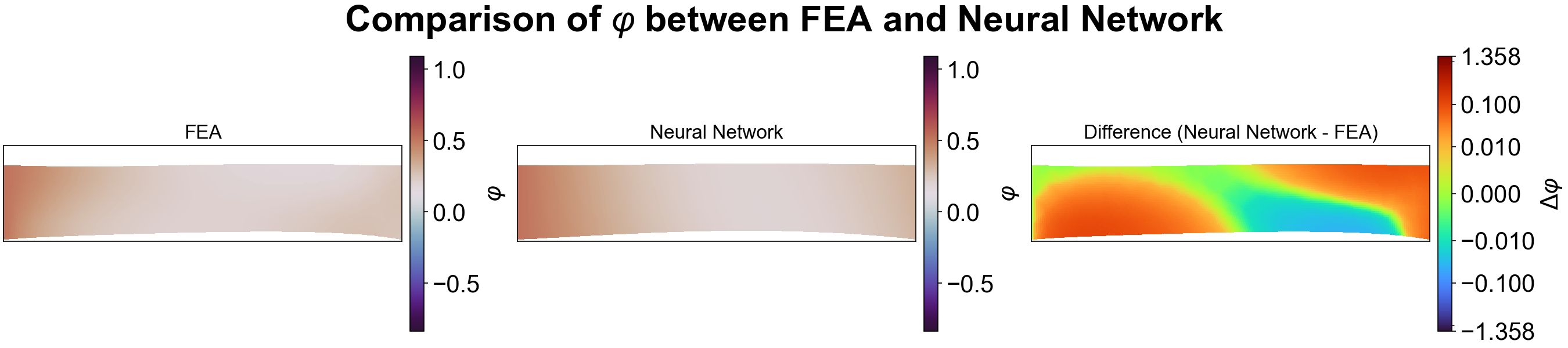}
        \caption{}
    \end{subfigure}

    \caption{Comparison between solutions obtained from FEA and NN for the case with $\varphi_0 = \pi/6$. From the right to the left, results from FEA, results from NN, and the difference between the solutions. (a) Magnitude of deformation with director fields, (b) Magnitude of deformation, (c) Magnitude of deformation in $x-$direction (d) Magnitude of deformation in $y-$direction, and (e) Orientation of the director field.}
    \label{pi6}
\end{figure}

For  $\varphi_0 = \pi / 6$ is shown in figure (\ref{pi6}), the director is initially closer to the loading direction. This leads to a deformation response that is slightly more anisotropic, with stronger coupling in the $X$ direction and subtle variations in $u_y$. In the final state, the FEA solution exhibits smooth gradients in both displacement components and a consistent director rotation toward the stretch direction.
The neural network again reproduces the displacement magnitude and component fields with strong agreement. Minor differences are visible near the boundaries and in regions where gradients are largest, but these differences remain small and smoothly distributed. The predicted director field converges to the same equilibrium pattern observed in the FEA simulation, demonstrating that the \textit{DirectorNet} successfully captures the coupling between deformation and orientational energy even when starting from a random initialization.

\subsubsection{Overall Assessment}

Across all three initial director orientations, the neural network accurately reproduces the final equilibrium displacement and director fields obtained from FEA. The largest discrepancies consistently occur in the primary loading direction ($X$), where gradients are strongest, while $u_y$ and $\varphi$ show very close agreement.
Crucially, the network is not trained on FEA data but instead minimizes the total potential energy directly. The agreement seen in the final states, therefore, indicates that the combined \textit{DeformationNet}--\textit{DirectorNet} architecture, together with the variational loss formulation, is sufficient to recover stable, energy--consistent equilibrium solutions of the coupled Cosserat model.

\section{Conclusion}
In this work, we developed energy--minimizing conditions for materials with microstructure by using theories of oriented media, also known as Cosserat elasticity, with a single unit director. We designed and implemented two multilayer perceptron networks to model the mechanical behavior of microstructured media by minimizing the total potential energy of these materials. This architecture guarantees that the outputs of the networks are kinematically independent and due to the specific form of enforcing the boundary conditions, these outputs are kinematically admissible too. We showed that the loss function satisfies the frame invariant conditions. In addition, we showed that the chosen constitutive model satisfies the strong form of the Legendre--Hadamard inequality, and as a result, the outputs are materially stable.  Furthermore, we validated the results of the network against the finite element method and the good agreement between the results verifies the output of the network.

\section{Codebase and Data Availability}
No data is used in this work. The codebase will be available by the corresponding author after publication. 

\section{Contribution}
MS: Writing and editing the original draft, formal analysis, theoretical development, neural network design and implementation, FEA design and implementation, network validation against FEA, and supervision. PHG: Writing and editing the original draft, neural network design, network validation against FEA, and visualization. MK: Network validation against FEA, and visualization. JLW: Writing and editing the original draft, mesh sensitivity analysis. FN: Writing and editing the original draft, and visualization.

\section{Acknowledgments}
The authors would like to thank Prof. Jay D. Humphrey at 
Yale University for his invaluable advice and guidance in the preparation of this work

\pagebreak
\printbibliography
\end{document}